\documentclass[sigconf]{acmart}

\usepackage{multirow}
\usepackage{enumitem}
\usepackage{tcolorbox}
\theoremstyle{acmplain}
\newtheorem{fairnessdefinition}{Definition}[section]
\newtheorem{fairnesstheorem}{Theorem}[section]
\tcbuselibrary{breakable}  
\tcbuselibrary{skins} 

\newcommand{\vpara}[1]{\vspace{0.05in}\noindent\textbf{#1 }}

\AtBeginDocument{%
  }

\setcopyright{acmlicensed}
\copyrightyear{2026}
\acmYear{2026}
\acmDOI{XXXXXXX.XXXXXXX}

\acmConference[This paper is under peer review]{}{}{}

\acmISBN{978-1-4503-XXXX-X/2018/06}

\begin{document}

\title{\texorpdfstring{PD$^3$}{PD3}: A Framework for Project Duplication Detection via Adapted Multi-Agent Debate}

\author{Dezheng Bao}
\email{baodezheng@zju.edu.cn}
\affiliation{%
  \institution{Zhejiang University}
  \city{Hangzhou}
  \state{Zhejiang}
  \country{China}
}

\author{Yueci Yang}
\email{yueciyang@zju.edu.cn}
\affiliation{%
  \institution{Zhejiang University}
  \city{Hangzhou}
  \state{Zhejiang}
  \country{China}
  }

\author{Chutian Yu}
\email{yu\_chutian@zj.sgcc.com.cn}
\affiliation{%
  \institution{State Grid Power Supply Co. Ltd.}
  \city{Hangzhou}
  \state{Zhejiang}
  \country{China}
  }

\author{Xin Chen}
\affiliation{%
  \institution{Zhejiang University}
  \city{Hangzhou}
  \state{Zhejiang}
  \country{China}
  }

\author{Zeguo Fei}
\affiliation{%
  \institution{Zhejiang University}
  \city{Hangzhou}
  \state{Zhejiang}
  \country{China}
  }

\author{Xiang Yuan}
\affiliation{%
  \institution{State Grid Power Supply Co. Ltd.}
  \city{Hangzhou}
  \state{Zhejiang}
  \country{China}
  }

\author{Lijun Zhang}
\affiliation{%
  \institution{State Grid Power Supply Co. Ltd.}
  \city{Hangzhou}
  \state{Zhejiang}
  \country{China}
  }

\author{Jiangqian Huang}
\affiliation{%
  \institution{State Grid Power Supply Co. Ltd.}
  \city{Hangzhou}
  \state{Zhejiang}
  \country{China}
  }

\author{Zhengxuan Jiang}
\affiliation{%
  \institution{Zhejiang University}
  \city{Hangzhou}
  \state{Zhejiang}
  \country{China}
  }

\author{Daoze Zhang}
\affiliation{%
  \institution{Zhejiang University}
  \city{Hangzhou}
  \state{Zhejiang}
  \country{China}
  }

\author{Junru Chen}
\affiliation{%
  \institution{Zhejiang University}
  \city{Hangzhou}
  \state{Zhejiang}
  \country{China}
  }

\author{Yang Yang}
\correspondingauthor
\email{yangya@zju.edu.cn}
\affiliation{%
  \institution{Zhejiang University}
  \city{Hangzhou}
  \state{Zhejiang}
  \country{China}
  }



\renewcommand{\shortauthors}{Bao et al.}

\begin{abstract}

  Project duplication detection is critical for project quality assessment because it helps avoid investment in repeated proposals. Existing methods usually cast it as ranking and rely on surface matching or direct large language models judging, often missing practical needs in set-level reference selection. We recast the task as many-to-many reference set selection, which requires broad candidate information and fair decomposed comparison under context limits. We propose \textbf{PD$^3$}, a framework for \textbf{P}roject \textbf{D}uplication \textbf{D}etection via adapted multi-agent \textbf{D}ebate. PD$^3$ combines local multi-agent debate with global round-robin scheduling to retrieve the relevant project set. Theoretically, this scheduler guarantees fair comparison through balanced exposure and comparison context. PD$^3$ also produces quantitative duplication scores and qualitative overlap feedback. On 800+ real-world power projects, PD$^3$ outperforms the strongest baselines by 4.05\% in relevant reference selection and 9.77\% in duplication score generation. We deploy \textbf{\textit{Review Dingdang}}, an online platform, which has helped save \$13.44 million across 442 new projects. Codes are available in \href{https://anonymous.4open.science/r/PD-3-KDD}{\textcolor{blue}{this repository}}.
\end{abstract}

\ccsdesc[500]{Computing methodologies~Natural language processing}
\ccsdesc[500]{Applied computing~Document management and text processing}



\maketitle

\section{Introduction}
\label{section-introduction}
Project duplication detection aims to assess how relevant a newly proposed project is to prior work~\citep{xiao2022oagbert, cohan-etal-2020-specter}. It retrieves relevant reference projects and provides evidence for expert review. This task is practically important for quality assessment and avoiding redundant research investments. Growing research investment makes this task increasingly difficult. For example, the State Grid Corporation of China (SGCC) invests 5.25 billion USD in science projects and reports 8,521 authorized patents~\citep{sgcc2023report}. The Smart Grid Grants program provides 300 billion USD in projects to improve the power system~\citep{doesmartgrid}. This expansion makes exhaustive manual detection costly and inconsistent. Moreover, both reviewers and applicants seek interpretable feedback rather than simple numerical results.


The primary objective is to retrieve the most relevant reference project set. Current retrieval pipelines usually cast this objective as ranking. They score candidates by relevance and take the top-$k$ as references. Word-frequency methods such as ROUGE~\citep{lin2004rouge} and BM25~\citep{robertson2009probabilistic} capture surface overlap, making them brittle to synonym substitution, paraphrasing, and intentional rewriting. Vector-distance methods, including embedding and reranking (e.g., gte, bge~\citep{chen2024bge,li2023making}), improve semantic matching, but compress long proposals into limited representations and often lose domain-specific evidence needed for final duplication decisions. They are therefore useful first-stage filters, but insufficient for fine-grained selection.

\begin{figure}[t]
  \centering
  \includegraphics[width=0.8\linewidth]{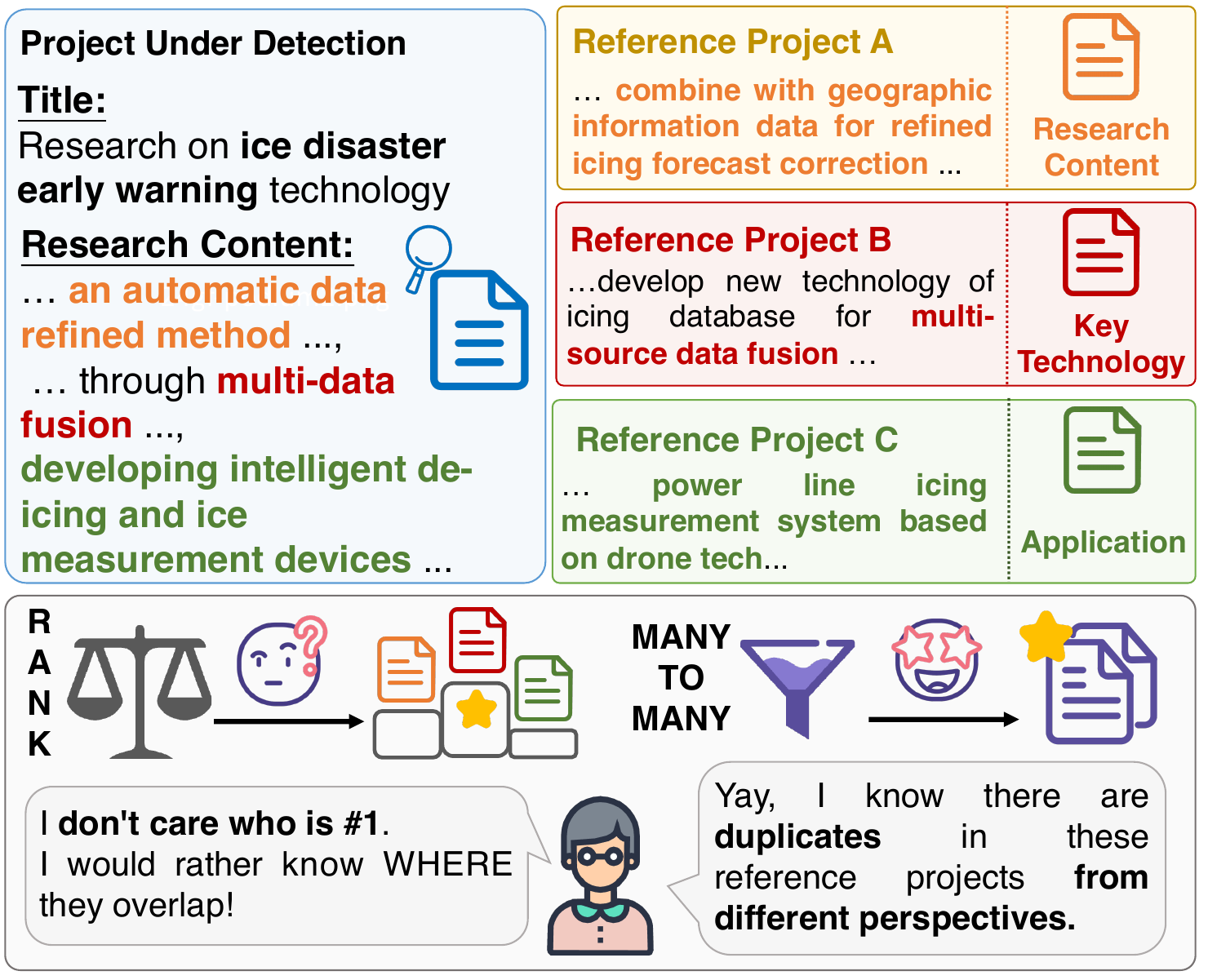}
  \vspace{-0.4cm}
  \caption{Project duplication retrieval scenario. Experts care more about where the overlaps occur from different perspectives than which reference project is ranked first.}
  \vspace{-0.4cm}
  \label{figure-set-selection}
\end{figure}

Large language model (LLM)-based methods can reason over richer project semantics. These methods have evolved through three broad paradigms: from single-LLM prompting and LLM-as-a-Judge scoring or pairwise comparison~\citep{zheng2023judging}, to rubric- or demonstration-guided calibration using explicit criteria or representative in-context demonstrations~\citep{shen2021mred, zhang2025re2consistencyensureddatasetfullstage}, and to Agent-as-a-Judge evaluation~\citep{chan2023chateval}. Among these, Multi-Agent Debate (MAD) is particularly appealing, as it resembles an expert committee review: multiple agents inspect complementary technical aspects, exchange evidence, and produce interpretable discussion traces~\citep{liang2023encouraging,du2023improving}. However, these paradigms remain difficult to deploy for project duplication detection. Detailed project-specific rubrics and representative demonstrations are scarce, and direct or single-LLM judgments are prone to bias and instability. More importantly, most methods still produce rankings rather than the evidence-bearing reference sets experts need.

From a scenario perspective, the first challenge is that \textbf{strict ranking is inadequate for project duplication retrieval}. Experts need a reference set that jointly supports evidence for duplication assessment, not an absolute order over all relevant historical projects. As shown in Figure~\ref{figure-set-selection}, a new proposal may overlap with different references in research content, technical route, or application scenario. A single scalar ranking is unnecessary and also difficult. Ranking can hide this multi-aspect evidence and cannot explain how references work together. We therefore reformulate retrieval as many-to-many reference set selection under expert-defined criteria.


From a task perspective, the second challenge is that \textbf{jointly comparing many candidates creates overlong LLM contexts}. A candidate's relevance depends not only on itself and the target proposal but also on other candidates, because their evidence can overlap or complement each other. Although preliminary retrieval removes many irrelevant projects, our analysis in Appendix~\ref{section-appendix-preliminary-retrieval} shows that covering most human-annotated references still requires dozens of candidates. Placing all long proposals in one prompt creates overlong contexts and lost-in-the-middle risks~\citep{liu2023lost}. This motivates decomposing set selection into smaller comparison groups while preserving broad global candidate information.

\begin{figure}[t]
  \centering
  \includegraphics[width=0.8\linewidth]{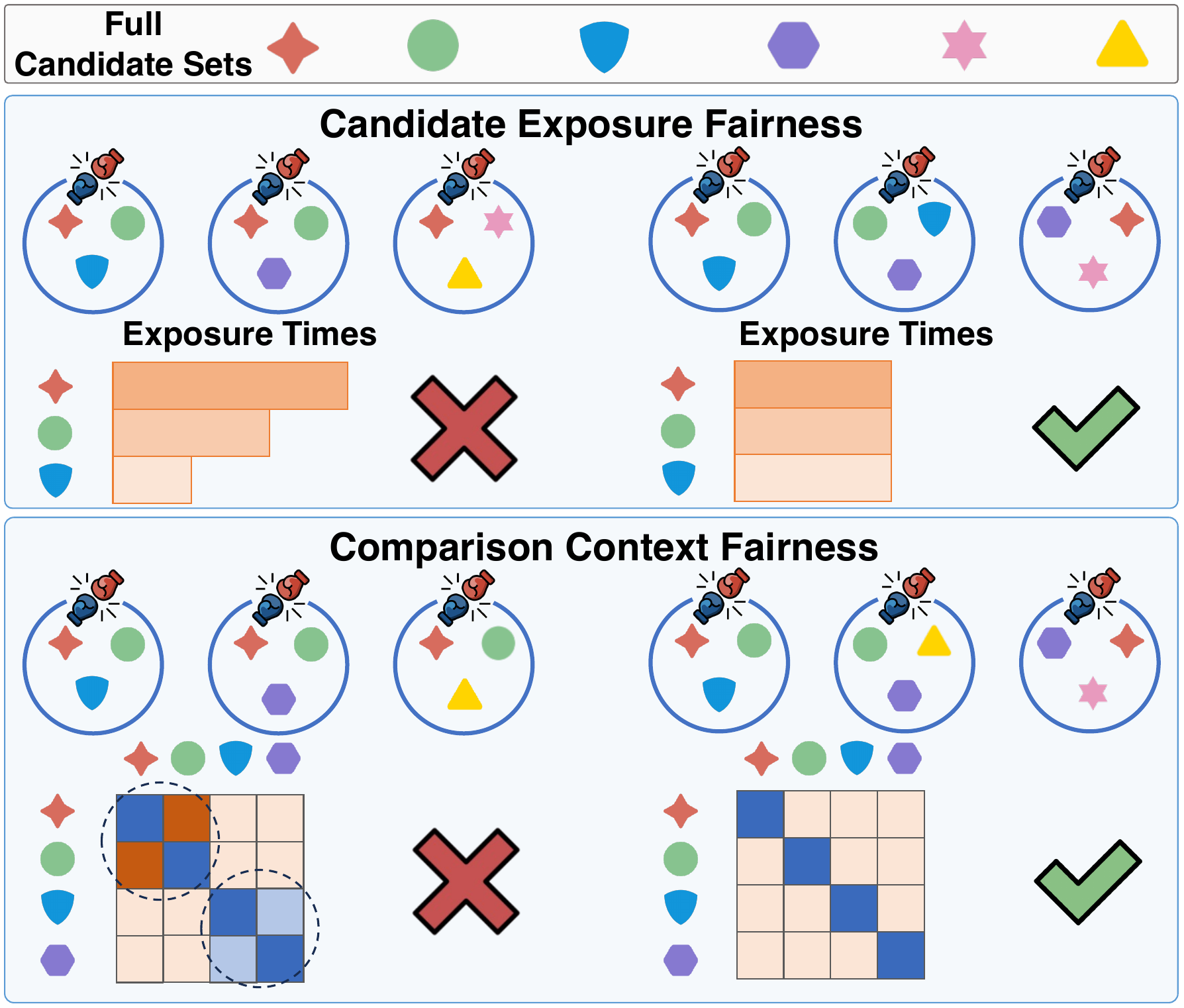}
  \vspace{-0.3cm}
  \caption{Two fairness requirements for decomposed candidate comparison. Candidate exposure fairness requires balanced evaluation opportunities. Comparison context fairness requires balanced co-evaluation contexts.}
  \label{figure-fairness}
  \vspace{-0.4cm}
\end{figure}


From a methodological perspective, the third challenge is that \textbf{decomposing candidates can introduce unfair comparisons}. If comparison groups are formed by simple partitioning or random sampling, some candidates may receive fewer evaluation opportunities or repeatedly appear in favorable or unfavorable contexts. As shown in Figure~\ref{figure-fairness}, we argue that fair decomposition should satisfy two requirements: \textbf{candidate exposure fairness}, where candidates receive balanced evaluation opportunities, and \textbf{comparison context fairness}, where candidates are evaluated with balanced and diverse groups of other candidates, so that no candidate is repeatedly compared with unusually weak, unusually strong, or overly homogeneous neighboring regions. Without these guarantees, a candidate may be selected because of its assigned comparison context rather than its globally strong duplication evidence.


To address these challenges, we propose \textbf{PD$^3$}, a framework for \textbf{P}roject \textbf{D}uplication \textbf{D}etection via adapted multi-agent \textbf{D}ebate. Rather than treating retrieval as strict ranking, PD$^3$ adapts MAD to select an evidence-bearing reference set through local expert-agent deliberation. To balance global information and context length constraint, PD$^3$ couples local debate with round-robin scheduling: it partitions candidates into blocks and enumerates block combinations as tractable comparison groups. Within each group, multiple expert agents independently select relevant references, debate with evidence, and a senior judge makes a group-level decision. PD$^3$ then aggregates group decisions by voting. We also formalize candidate exposure and comparison-context fairness and propose theoretical prove for comparison unbias of PD$^3$. Beyond retrieval, PD$^3$ produces reviewer-facing feedback: a quantitative agent assigns a duplication score, while qualitative agents summarize the main overlaps and highlight semantically similar text segments.

We further deploy PD$^3$ in \textbf{\textit{Review Dingdang}}, an online detection platform for power systems. In 2025, the platform reviewed 442 newly proposed projects and prevented 13.44 million USD from being invested in duplicate projects. Our key contributions are:

\vspace{-0.1cm}
\begin{itemize}[leftmargin=0.35cm]

    \item We reformulate project duplication retrieval as many-to-many relevant reference set selection rather than strict ranking, better matching the expert-review need.

    \item We come up with two fairness criteria for project duplication detection, which take into account key factors in the comparison of reference project candidates.

    \item We propose \textbf{PD$^3$}, a novel adapted multi-agent debate framework that balances global candidate coverage with LLM context-length constraints. With a novel round-robin scheduling strategy, PD$^3$ balances candidate exposure and co-evaluation contexts, and also provides reviewer-facing feedback. We also provide a theoretical analysis of the fairness guarantees of PD$^3$

    \item We validate PD$^3$ on real-world and cross-domain datasets and deploy the framework in an online platform, \textbf{\textit{Review Dingdang}}, demonstrating its practical effectiveness.

\end{itemize}

\begin{figure*}
  \centering
  \includegraphics[width=\textwidth]{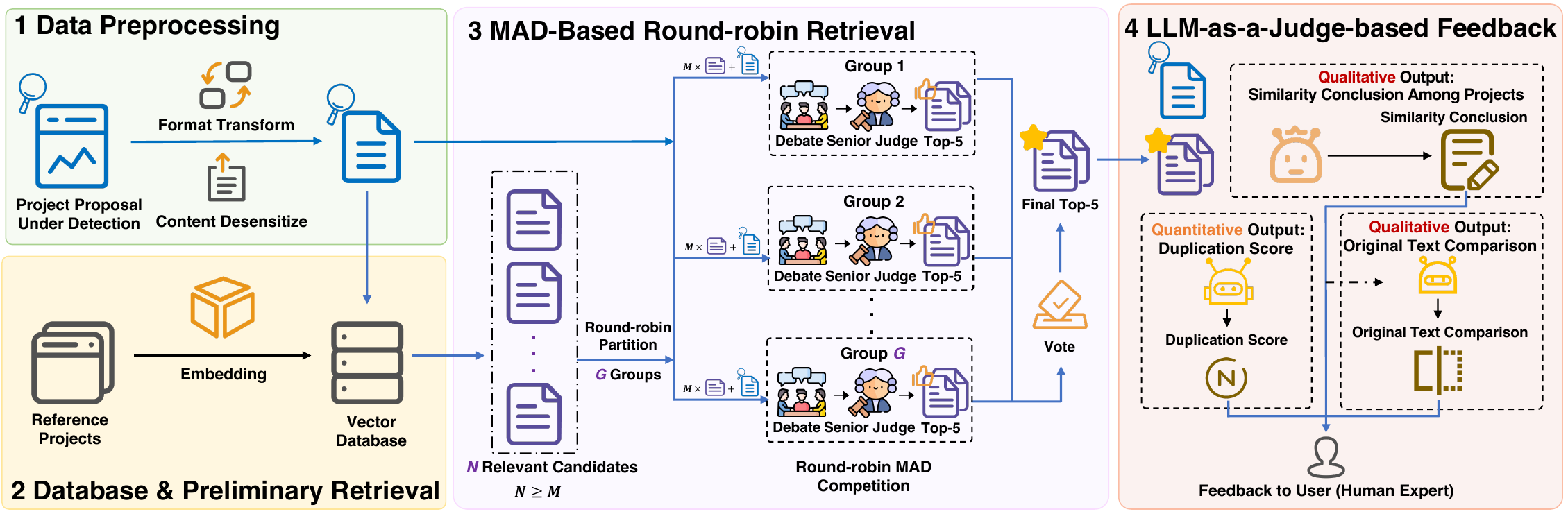}
  \vspace{-0.2cm}
  \caption{Overview of PD$^3$ framework. The detection pipeline consists of four modules: data pre-processing, database and preliminary retrieval, MAD-based round-robin retrieval, and LLM-as-a-Judge-based Feedback.}
  \label{figure-framework}
  \vspace{-0.4cm}
\end{figure*}

\section{Preliminary}
\label{section-preliminary}

\subsection{Notation and Task Definition}
\label{section-preliminary-task}

We first formalize project duplication detection and the notation used throughout the paper. Let $P_u$ denote the project under detection, and let $\mathcal{S}=\{P_i\}_{i=1}^{L}$ denote the historical reference project corpus. Since directly comparing $P_u$ with all historical projects is computationally expensive, a preliminary retriever first returns a candidate pool $\mathcal{V}=\{v_i\}_{i=1}^{N}\subseteq\mathcal{S}$, where each $v_i$ is a candidate reference project. Let $\mathcal{Q}$ denote the expert-defined duplication-detection criteria, such as research content, key technology, and application scenario. We use $k$ to denote the number of final reference projects selected for downstream detection, and use $\mathcal{R}\subseteq\mathcal{V}$ with $|\mathcal{R}|=k$ to denote the selected relevant reference set.

\vpara{Task 1: Relevant Reference Set Selection.} The first task is to select a relevant reference set $\mathcal{R}$ for $P_u$ from the candidate pool $\mathcal{V}$. Different from ranking-oriented retrieval, this task does not require a strict total order over all candidates. Instead, the goal is to identify a compact set of historical projects that jointly provide sufficient duplication evidence under $\mathcal{Q}$:

\begin{equation}
    \label{eq:task1}
    \mathcal{R}=F(P_u,\mathcal{V},\mathcal{Q}),\qquad
    \mathcal{R}\subseteq\mathcal{V},\quad |\mathcal{R}|=k.
\end{equation}

This formulation better matches project duplication detection because different reference projects may overlap with $P_u$ from different aspects. The selected set therefore emphasizes evidence coverage rather than an absolute ranking among candidates.

\vpara{Task 2: Duplication Score Generation.} The second task is to produce quantitative detection feedback based on $P_u$ and the selected reference set $\mathcal{R}$. Let $\mathcal{D}$ denote the score domain. The quantitative component should assign a project-level overall duplication score:

\begin{equation}
    \label{eq:task2}
    s_u=f(P_u,\mathcal{R},\mathcal{Q}),\qquad s_u\in\mathcal{D}.
\end{equation}


\subsection{Comparison Fairness}
\label{section-preliminary-fairness}

Many-to-many set selection requires broad candidate information, but LLM agents cannot reliably evaluate all $N$ candidates in a single context when projects are long. We therefore define a candidate comparison scheduler $\mathcal{A}$ that decomposes the candidate pool into a collection of context-bounded comparison groups:
\begin{equation}
    \label{eq:scheduler}
    \mathcal{C}=\mathcal{A}(\mathcal{V})
    =\{C_g\}_{g=1}^{G},\qquad
    C_g\subseteq\mathcal{V},\quad |C_g|=M<N.
\end{equation}

Each $C_g$ is a \textbf{comparison context} in which LLM agents jointly evaluate $M$ candidates. Because the scheduler determines which candidates are jointly evaluated, it directly affects whether the final set selection is fair. We then define two kinds of fairness because project relevance is context-dependent: a candidate's chance of being selected depends not only on its own overlap with $P_u$, but also on the other candidates placed in the same comparison context. 

\vpara{Candidate Exposure Fairness.} Candidate exposure fairness requires every candidate to receive balanced evaluation opportunities.

\begin{fairnessdefinition}[Candidate Exposure Fairness]\label{def:candidate-exposure}
For each candidate $v_i$, define its group-inclusion count and exposure gap as
\begin{equation}
    \label{eq:exposure_times}
    a_i=\sum_{g=1}^{G}\mathbb{I}(v_i\in C_g).
\end{equation}
\begin{equation}
    \label{eq:exposure_fairness}
    \Delta_{\mathrm{incl}}=\max_i a_i-\min_i a_i,\qquad \Delta_{\mathrm{incl}} \ge 0.
\end{equation}
\end{fairnessdefinition}

When all groups share the same size, the scheduler satisfies candidate exposure fairness if and only if $\Delta_{\mathrm{incl}}=0$; smaller values indicate a more balanced schedule.

\vpara{Comparison Context Fairness.} Comparison context fairness requires candidates to be evaluated in balanced and diverse contexts. We characterize this requirement from both pair-level and group-level perspectives.

\begin{fairnessdefinition}[Pair-level Comparison Context Fairness]\label{thm:pair-level-comparison-context}
For each candidate pair $(v_i,v_j)$, define its co-evaluation count as
\begin{equation}
    \label{eq:co-eval_times}
    c_{ij}=\sum_{g=1}^{G}\mathbb{I}(\{v_i,v_j\}\subseteq C_g), \qquad \gamma_{\mathrm{pair}}=\min_{i<j}c_{ij}.
\end{equation}

For a target coverage threshold $\tau$, the number of insufficiently covered pairs is
\begin{equation}
    \label{eq:low_coverage_pair_count}
    L_{\mathrm{pair}}(\tau)=\sum_{i<j}\mathbb{I}(c_{ij}<\tau),
\end{equation}
\end{fairnessdefinition}

Full pairwise coverage holds if and only if $\gamma_{\mathrm{pair}}\ge 1$. Every pair is jointly evaluated at least $\tau$ times if and only if $L_{\mathrm{pair}}(\tau)=0$.

\begin{fairnessdefinition}[Group-level Comparison Context Fairness]\label{thm:group-level-comparison-context}
For any two contexts $C_g$ and $C_h$, define their overlap as
\begin{equation}
    \label{eq:context_overlap}
    o_{gh}=|C_g\cap C_h|,\qquad
    o_{\max}=\max_{g<h}o_{gh}.
\end{equation}

The number of highly redundant context pairs under threshold $q$ is
\begin{equation}
    \label{eq:highly_redundant_pairs}
    R_{\mathrm{ctx}}(q)=\sum_{g<h}\mathbb{I}(o_{gh}>q).
\end{equation}
\end{fairnessdefinition}
Lower $o_{\max}$ and $R_{\mathrm{ctx}}(q)$ indicate more diverse comparison contexts. Together with high pairwise coverage, group-level metrics further characterize comparison context fairness.

\section{Methods: PD$^3$ Framework}
\label{section-frameworkdesign}

\subsection{Overview}
\label{section-frameworkdesign-overview}

As illustrated in Figure~\ref{figure-framework}, PD$^3$ first builds a clean and searchable candidate pool before invoking LLM-based reasoning. \textbf{(1) Data Pre-processing.} The raw project files may contain heterogeneous formats and sensitive information. PD$^3$ standardizes these inputs by unifying file formats, anonymizing sensitive expressions with rule-based filters, and extracting textual project content. \textbf{(2) Database and Preliminary Retrieval.} PD$^3$ maintains an offline-updatable reference project database and uses vector retrieval to obtain the candidate pool $\mathcal{V}$ for the project under detection $P_u$. These two modules provide a tractable candidate set, while the following round-robin MAD retrieval and LLM-as-a-Judge feedback modules perform set selection and feedback generation.

\subsection{Round-robin MAD Retrieval: Fair Set Selection under Context Constraints}
\label{section-frameworkdesign-mad}
Accurately selecting relevant reference projects is the most critical component of project duplication detection, because retrieval errors directly affect downstream scoring and evidence generation. Following the scenario insights in Section~\ref{section-introduction} and the task formulation in Section~\ref{section-preliminary-task}, we model retrieval as many-to-many set selection and propose an adapted MAD mechanism.

Round-robin MAD retrieval addresses the two challenges identified above: preserving global candidate information under LLM context-length constraints and ensuring fair comparison scheduling after decomposition. Directly applying MAD to all $N$ candidates would overload the context with long project texts and multi-round discussion records, whereas evaluating only a small subset would lose global candidate coverage. Inspired by practical expert discussion, PD$^3$ therefore decomposes $\mathcal{V}$ into multiple comparison groups, runs local MAD-based selection in each group, and aggregates group-level decisions into the final reference set $\mathcal{R}$.

\vpara{Round-robin scheduler.} Given the candidate pool $\mathcal{V}=\{v_i\}_{i=1}^{N}$, PD$^3$ instantiates the scheduler $\mathcal{A}$ in Eq.~\ref{eq:scheduler} with a round-robin block design. Assuming $N$ is divisible by the block size $B$, we partition $\mathcal{V}$ into $J=N/B$ disjoint blocks: $\mathcal{B}_1,\mathcal{B}_2,\ldots,\mathcal{B}_J,\qquad |\mathcal{B}_j|=B.$


Each comparison group combines \(r=M/B\) blocks. By enumerating all non-repeating block combinations, PD$^3$ constructs \(\displaystyle G=\binom{J}{r}\) comparison groups. This design can be viewed as a tractable block-level approximation to exhaustive candidate-level comparison: exhaustive enumeration produces $\binom{N}{M}$ groups, whereas round-robin requires only $\binom{N/B}{M/B}$ groups.

\vpara{MAD-based group evaluation and voting aggregation.} Each comparison group $C_g$ is evaluated independently by multiple expert agents under the criteria $\mathcal{Q}$. In the initial round, each agent selects $k$ candidate references from $C_g$ and provides evidence. In the debate rounds, agents inspect previous arguments, critique conflicting choices, and revise their selections when stronger evidence is provided. A senior expert agent then reads the full discussion record and outputs a local selected set $\mathcal{R}_g\subseteq C_g$ with $|\mathcal{R}_g|=k$. After all groups finish, PD$^3$ aggregates local decisions by voting: \(\displaystyle vote_i=\sum_{g=1}^{G}\mathbb{I}(v_i\in\mathcal{R}_g).\) The final selected reference set $\mathcal{R}$ consists of the $k$ candidates with the largest vote counts.


Since comparison groups are independent, they can run in parallel, and the voting step integrates local evidence into a global set-selection result. This aggregation is also consistent with prior findings that voting is effective for knowledge-based tasks~\citep{kaesberg2025voting}.

\vpara{Fairness Analysis of Round-robin Scheduling.} Round-robin scheduling provides deterministic guarantees for the fairness metrics defined in Section~\ref{section-preliminary-fairness}. We formalize these guarantees below.

\begin{fairnesstheorem}[Uniform candidate exposure]
\label{thm:rr_uniform_exposure}
For the round-robin block scheduler with $J=N/B$ blocks and $1\le r=M/B\le J$ blocks per comparison group, every candidate is evaluated in the same number of comparison groups:
\begin{equation}
    \label{eq:rr_exposure}
    a_i^{\mathrm{RR}}=\binom{J-1}{r-1},\qquad \forall i.
\end{equation}
Consequently, the inclusion disparity is zero, i.e.,
\begin{equation}
    \Delta_{\mathrm{incl}}^{\mathrm{RR}}=0.
\end{equation}
\end{fairnesstheorem}

\begin{proof}
Let $b(i)$ denote the block index of candidate $v_i$. Candidate $v_i$ is included in a comparison group if and only if its block $\mathcal{B}_{b(i)}$ is selected. Once this block is fixed, the remaining $r-1$ blocks can be chosen from the other $J-1$ blocks in $\binom{J-1}{r-1}$ ways. This count does not depend on $i$, so all candidates receive the same number of evaluation opportunities and $\Delta_{\mathrm{incl}}^{\mathrm{RR}}=0$.
\end{proof}


    

\begin{fairnesstheorem}[Pairwise co-evaluation coverage]
\label{thm:rr_pairwise_coverage}
For any pair of candidates $(v_i,v_j)$, the round-robin block scheduler with $2\le r\le J$ gives the following co-evaluation count:
\begin{equation}
    \label{eq:rr_pair_coevaluation}
    c_{ij}^{\mathrm{RR}}=
    \begin{cases}
    \binom{J-1}{r-1}, & b(i)=b(j),\\
    \binom{J-2}{r-2}, & b(i)\ne b(j).
    \end{cases}
\end{equation}
Thus, when $2\le r\le J$, the minimum pairwise co-evaluation count is
\begin{equation}
    \label{eq:rr_gamma_pair}
    \gamma_{\mathrm{pair}}^{\mathrm{RR}}
    =
    \min_{i<j}c_{ij}^{\mathrm{RR}}
    =
    \binom{J-2}{r-2}\ge 1.
\end{equation}
\end{fairnesstheorem}

\begin{proof}
If $v_i$ and $v_j$ belong to the same block, they are jointly evaluated whenever that block is selected. The remaining $r-1$ blocks can be chosen from the other $J-1$ blocks, yielding $\binom{J-1}{r-1}$ co-evaluations. If they belong to different blocks, both corresponding blocks must be selected, and the remaining $r-2$ blocks are chosen from the other $J-2$ blocks, yielding $\binom{J-2}{r-2}$ co-evaluations. Since $\binom{J-1}{r-1}\ge \binom{J-2}{r-2}$ for $2\le r\le J$, the minimum pairwise count is $\binom{J-2}{r-2}$. This value is at least one, so every reference project candidate pair is jointly evaluated at least once.
\end{proof}

\begin{fairnesstheorem}[Bounded context redundancy]
\label{thm:rr_context_redundancy}
For the round-robin block scheduler with $1\le r<J$, any two distinct comparison groups have candidate-level overlap bounded by
\begin{equation}
    \label{eq:rr_context_overlap}
    o_{\max}^{\mathrm{RR}}=B(r-1)=M-B.
\end{equation}
Therefore, no two comparison contexts of all groups overlap by more than $M-B$ candidates.
\end{fairnesstheorem}

\begin{proof}
Each comparison group corresponds to a distinct subset of $r$ blocks. If two distinct groups shared all $r$ blocks, they would be the same group, contradicting distinctness. Hence two distinct groups can share at most $r-1$ blocks. Since blocks are disjoint and each block contains $B$ candidates, their candidate-level overlap is at most $B(r-1)$. Using $M=Br$, this bound equals $M-B$.
\end{proof}

\vpara{Instantiation Analysis.} Based on candidate-pool recall, LLM context capacity, and scheduling efficiency (details in Appendix~\ref{section-appendix-expsettings}), we set $N=30$, $M=20$, and $B=5$. This gives $J=6$ blocks, $r=4$ blocks per comparison context, and $G=\binom{6}{4}=15$ contexts. With a fixed per-context MAD configuration, total scheduler work is $O(N+G)$, with group evaluation scaling with $G$, whereas exhaustive candidate-level enumeration scales as $O\!\left(\binom{N}{M}\right)$. In the main setting, round-robin therefore evaluates 15 contexts instead of $\binom{30}{20}=30{,}045{,}015$ contexts, a reduction by a factor of 2{,}003{,}001.

The same construction provides deterministic fairness guarantees. Every candidate appears in $\binom{5}{3}=10$ contexts, so $\Delta_{\mathrm{incl}}^{\mathrm{RR}}=0$. Same-block pairs are co-evaluated 10 times and cross-block pairs 6 times, yielding $\gamma_{\mathrm{pair}}^{\mathrm{RR}}=6$ and $L_{\mathrm{pair}}^{\mathrm{RR}}(6)=0$. At the group level, $o_{\max}^{\mathrm{RR}}=15$, so $R_{\mathrm{ctx}}^{\mathrm{RR}}(q)=0$ for $q\ge15$.

\subsection{LLM-as-a-Judge Based Feedback: Quantitative and Qualitative Output}
\label{section-frameworkdesign-feedback}
Through literature review and expert consultation, we identify the lack of constructive feedback as another limitation. Effective feedback should bridge human-system interaction with actionable insights to improve detection efficiency and project quality. 

To address this, we develop an LLM-based agent module that delivers comprehensive quantitative and qualitative feedback. After selecting $\mathcal{R}$, PD$^3$ generates detection feedback with an LLM-as-a-Judge module. The quantitative agent takes $P_u$, $\mathcal{R}$, and the criteria $\mathcal{Q}$ as input, and assigns a duplication score $s_u\in\mathcal{D}$ using a predefined scoring scale. Compared with point-wise scoring over individual references, this set-level scoring process evaluates the project under detection against the whole selected reference set, which better matches the many-to-many setting.

PD$^3$ also produces concise qualitative evidence for human experts. The \textbf{similarity-conclusion} agent summarizes the main overlaps between $P_u$ and $\mathcal{R}$ in terms of research content, key technologies, and application scenarios. The \textbf{original-text-comparison} agent further identifies semantically similar text segments from the source projects. These outputs make the final score easier to inspect and help experts locate the concrete project contents that support the duplication detection result.

\begin{table*}
  \caption{Experiment results of Task 1: Most relevant top-5 retrieval.}
  \vspace{-0.3cm}
  \label{retrieval-exp-results}
  \centering
  \footnotesize
  \resizebox{0.85\textwidth}{!}{
    \begin{tabular}{crrrrrr}
        \toprule
        \multirow{3}{*}{\textbf{Method}} & \multirow{3}{*}{\textbf{Prec@5}}  & \multicolumn{5}{c}{\textbf{Match@K}} \\
        \cmidrule{3-7}
        &  & \textbf{K = 1} & \multirow{2}{*}{\textbf{K = 2}} & \multirow{2}{*}{\textbf{K = 3}} & \multirow{2}{*}{\textbf{K = 4}} & \multirow{2}{*}{\textbf{K = 5}}\\
        & & \textbf{(Hit Rate@5)} & & & &  \\
        \midrule
        Random \textbf{(Random)}  & 16.80 & 209 | 63.14 & 64 | 19.34 & 5 | 1.51 & 0 | 0.00 & 0 | 0.00 \\
        \cmidrule{1-7}
        ROUGE-L \textbf{(WF)} & 23.99  & 248 | 74.92 & 125 | 37.76 & 23 | 6.95 & 1 | 0.30 & 0 | 0.00  \\
        BM25 \textbf{(WF)}  & 28.70 & 273 | 82.48 & 145 | 43.81 & 52 | 15.71 & 5 | 1.51 & 0 | 0.00 \\
        \cmidrule{1-7}
        gte-1.5B \textbf{(VD)} & 38.13 & 296 | 89.43  & 219 | 61.66 & 97 | 29.31 & 18 | 5.44 & 1 | 0.30  \\
        gte-1.5B with instruction \textbf{(VD)} & 37.82 & 294 | 88.82 & 213 | 64.35 & 96 | 29.00 & 21 | 6.36 & 2 | 0.60  \\
        gte-7B as reranker \textbf{(VD)} & 38.97  & 298 | 90.03 & 212 | 64.05 & 105 | 31.72 & 29 | 8.76 & 1 | 0.30  \\
        Qwen3-8B as reranker \textbf{(VD)}~\citep{qwen3embedding}& \underline{40.18}  & 301 | 90.94 & 220 | 66.47 & 110 | 33.23 & 32 | 9.67 & 2 | 0.60  \\
        jina as reranker \textbf{(VD)}~\citep{sturua2024jina}& 27.98  & 275 | 83.08 & 141 | 42.60 & 42 | 12.69 & 4 | 1.21  & 1 | 0.30  \\
        \cmidrule{1-7}
        DeepSeek V3 \textbf{(LLM)}~\citep{deepseekai2024deepseekv3technicalreport}&  36.86 & 297 | 89.73 & 194 | 58.61 & 88 | 26.59 & 27 | 8.16 & \underline{4 | 1.21}  \\
        DeepSeek R1 \textbf{(LLM)}~\citep{guo2025deepseek}& 39.52 & 298 | 90.03 & 214 | 64.65  & 110 | 33.23 & 29 | 8.76 & 3 | 0.91  \\
        LLM-as-a-Judge \textbf{(LLM)} & 38.49  & 303 | 91.54 & 211 | 63.75 & 99 | 29.91 & 21 | 6.34 & 3 | 0.91 \\
        TourRank \textbf{(LLM)}~\citep{chen2025tourrank}& 37.04  & 290 | 87.61 & 193 | 58.31 & 100 | 30.21 & 27 | 8.16 & 3 | 0.91 \\
        \cmidrule{1-7}
        MAD Vanilla \textbf{(MAD)} & 39.64  & \underline{307 | 92.75} & \underline{229 | 69.18} & \underline{129 | 38.97} & \textbf{42 | 12.69} &  3 | 0.91  \\
        DMAD \textbf{(MAD)}~\citep{liu2025breaking}& 38.85  & 295 | 89.12 & 215 | 64.95 & 102 | 30.82 & 31 | 9.37 &  0 | 0.00  \\
        \cmidrule{1-7}
        \textbf{PD$^3$ MAD Round-robin (Ours)}  & \textbf{44.23}  & \textbf{310 | 93.66} & \textbf{238 | 71.90}  & \textbf{133 | 40.18} & \underline{41 | 12.39}  & \textbf{10 | 3.02}  \\
        \bottomrule
      \end{tabular}
    }
\end{table*}

\section{Experiments}
\label{section-exp}

In this section, we present a comprehensive evaluation of PD$^3$. Our investigation is guided by the following research questions: \textbf{RQ1}: How effective is PD$^3$'s retrieval? \textbf{RQ2}: How useful is PD$^3$ feedback? \textbf{RQ3}: How does round-robin scheduling affect comparison fairness and retrieval performance? \textbf{RQ4}: How well does PD$^3$ perform across domains? \textbf{RQ5}: How robust is PD$^3$ to its hyperparameters?

We address RQ1 and RQ2 through primary experimental analysis in Section~\ref{section-exp-results}, and discuss RQ3 and RQ4 via fairness-oriented ablation study and cross-domain analysis in Section~\ref{section-exp-ablation} and Section~\ref{section-exp-cross-domain}. For RQ5, we conduct detailed analysis on base models, the number of candidates in each comparison group, and the number of agents or debate rounds in Section~\ref{section-exp-hyperparameter}.

\subsection{Dataset}
\label{section-exp-dataset}

To rigorously evaluate PD$^3$'s real-world performance, we use and analyze a dataset of 833 scientific and technological projects across 3 years (from 2022 to 2024) from SGCC for validation, with an average raw data length of ~27.1k words and ~14.6k tokens.

These projects focus on new technologies in power systems and cross-application of cutting-edge technologies in other fields, such as artificial intelligence. Common topics include AI-based power consumption forecasting, line icing prediction, and carbon emission detection. Spanning 22 distinct domains (e.g., dispatching, digitalization, and informatization), the dataset captures the breadth of modern power systems. A single project often involves both real-world power-system scenarios and technical solutions from other fields, further increasing the complexity of the duplication detection. See details in Appendix~\ref{section-appendix-dataset}.

\subsection{Experiment Settings}
\label{section-exp-setting}

\vpara{Tasks.} We evaluate PD$^3$ on the two tasks defined in Section~\ref{section-preliminary-task}: \textit{Relevant Reference Set Selection} and \textit{Duplication Score Generation}. In our experiments, Task 1 uses preliminary retrieval to form a 30-project candidate pool and asks each method to select the top-5 relevant reference projects, while Task 2 uses the selected reference set to produce a comprehensive duplication score.

For Task 1, we randomly select 331 projects as test items to control annotation cost. Human experts annotate the optimal top-5 reference set $\hat{R} = \{\hat{P}_j^*\}_{j=1}^5$ for each test item. We employ a cross-validation style detection setting: when one project is under detection, all other 832 projects serve as reference candidates. This setting is closer to the cross-checking required for the same batch of projects in the real world.

For Task 2, different algorithms may use distinct score ranges $\mathcal{D}$ and scoring functions $f: (P_{\text{u}}, R) \rightarrow \mathcal{D}$, so we convert score evaluation into pairwise comparison for fair assessment. We randomly sample 256 project pairs $C = \{(P_{\text{u-A}j},P_{\text{u-B}j})\}_{j=1}^{256}$ among the 331 human-annotated projects. Three experts independently vote on which project has higher duplication in each pair, yielding annotations $\hat{H} = \{H_j\}_{j=1}^{256}$, where $H_j \in \{\text{u-A}, \text{u-B}\}$.

\vpara{Baselines.} For a comprehensive evaluation, we compare our approach against baselines from four categories: (1) Word frequency-based (\textbf{WF}) methods (e.g., ROUGE-L, BM25); (2) Vector distance-based (\textbf{VD}) models, including lightweight embedders (gte-1.5B) and rerankers (e.g., Qwen3-8B, jina, bge); (3) LLM-based (\textbf{LLM}) approaches, including strong models such as DeepSeek V3/R1 and paradigms such as LLM-as-a-Judge and TourRank; and (4) MAD-based (\textbf{MAD}) methods, including Vanilla MAD and DMAD. Detailed descriptions of all baselines are provided in Appendix~\ref{section-appendix-expsettings}.

\vpara{Settings} In Task 1, we employ gte-Qwen2-1.5B-instruction (without instruction) as the embedder for preliminary retrieval to obtain 30 candidate projects. In Task 2, for baseline methods that do not directly output a duplication score, we report both the maximum and average score among top-5 candidates for fair comparison. To comprehensively evaluate Task 2, we implement two settings using different top-5 reference sets ($R$): \textbf{(1) Controlled (Fixed Candidates):} Uniform expert-annotated reference set for direct scoring method comparison. \textbf{(2) End-to-End (Self-Retrieved Candidates):} Each method utilizes its independently retrieved reference set to assess overall end-to-end detection capabilities. See details in Appendix~\ref{section-appendix-expsettings}.

\vpara{Evaluation Metrics} In Task 1, we adopt 2 metrics: \textbf{Precision@5 (Prec@5)} and \textbf{Match@K}. $\text{Precision@5}=|R \cap \hat{R}| / 5$ measures selection overlap with experts. $\text{Match@K} = \sum \mathbb{I} ( \big| R \cap \hat{R} \big| \geq K ), K = 1,2,\dots, 5$ calculates the results that overlap with the expert selection greater than or equal to K, where $\mathbb{I}(\cdot)$ is the indicator function (1 if true, else 0). Particularly, $\text{Match@1} \equiv \text{Hit Rate@5}$. For $\text{Match@K}$, we additionally report its ratio to the size of test set in the form $\text{Match@K | } (\text{Match@K}/\text{Size of test set}$). In Task 2, we use \textbf{Accuracy (Acc)} on 256 test sets, with 2 evaluation approaches: \textbf{(1) ACC}: Strict expert majority as ground truth. \textbf{(2) Weighted ACC}: Expert votes as weights (e.g., 2A:1B $\rightarrow$ B scores 0.33). We incorporate this weighted setting because expert disagreement often reflects the inherent difficulty of the comparison and the multi-dimensional nature of the analysis. We report results in percentages for clarity.

\subsection{Experiment Results}
\label{section-exp-results}

\vpara{Analysis of Task 1 (RQ1)} Table~\ref{retrieval-exp-results} presents the experimental results for Task 1\footnote{Only the stronger VD baselines are shown. Full results are in Table~\ref{retrieval-exp-results-other}.}. PD$^3$ achieves the best Precision@5 of 44.23 and its improvement over each baseline is statistically significant on Precision@5 (T-test results are provided in Table~\ref{tab:statistical_test_task1}). Compared with the strongest baseline, Qwen3-8B as reranker, PD$^3$ improves Precision@5 by 4.05 percentage points. This comparison is important because vector-distance and reranking methods are suitable for preliminary retrieval, but their lower final-set accuracy confirms the limitation discussed in Section~\ref{section-introduction}: compressed similarity representations often miss the fine-grained, criterion-specific evidence needed for final duplication assessment.

The baseline patterns further show why PD$^3$ needs both multi-agent reasoning and fair context-bounded decomposition. WF methods remain far below PD$^3$ (23.99--28.70 Precision@5), indicating that surface overlap is insufficient for long technical projects with paraphrased or multi-aspect similarity. Direct LLM-based methods also do not close the gap: the best direct LLM result reaches 39.52 Precision@5, and TourRank reaches only 37.04, suggesting that applying a strong model or repeated tournament-style filtering does not by itself solve many-to-many reference set selection. Standard MAD variants likewise remain below PD$^3$: PD$^3$ improves over MAD Vanilla and DMAD by 4.59 and 5.38 percentage points, respectively. This supports our central design choice: debate is useful only when it is adapted to the long-context setting through balanced comparison groups and global voting aggregation.

The Match@K results provide a stricter view of set-level retrieval quality, while a higher $K$ value indicates increased task difficulty. PD$^3$ obtains the best or near-best results in all $K$ settings, recovering at least one expert-selected reference in 310 of 331 cases and recovering all five expert-selected references in 10 cases. The strong Match@5 result is especially relevant to our task formulation: compared with the strongest baseline on this metric (4 cases), PD$^3$ more than doubles the number of exact top-5 recoveries. This indicates that round-robin MAD is not merely finding isolated relevant candidates, but is better at constructing a complete evidence-bearing reference set. This also highlights that our method is more competitive and has greater application value in complex scenarios.

\begin{table}[h]
  \caption{Experiment results of Task 2.}
  \vspace{-0.3cm}
  \label{scoring-exp-results}
  \centering
  \resizebox{\linewidth}{!}{
      \begin{tabular}{crrrr}
        \toprule
        \multirow{2}{*}{\textbf{Method}} & \multicolumn{2}{c}{\textbf{Controlled}} & \multicolumn{2}{c}{\textbf{End-to-End}} \\
         & \textbf{ACC}  & \textbf{Weighted ACC} & \textbf{ACC}  & \textbf{Weighted ACC} \\
        \midrule
        ROUGE-L MAX \textbf{(WF)} & 55.47 & 55.34 & 53.91 & 54.04  \\
        ROUGE-L AVG \textbf{(WF)}  & 55.47 & 56.12 & 55.47 & 55.34 \\
        \cmidrule{1-5}
        BM25 MAX  \textbf{(WF)} & 56.64 & 56.77 & \underline{58.59} & 57.42 \\
        BM25 AVG  \textbf{(WF)} & 55.86 & 55.73  & 57.42 & 56.77 \\
        \cmidrule{1-5}
        gte-1.5B MAX  \textbf{(VD)}  & \underline{61.33} & 41.41 & 57.03 & 46.48 \\
        gte-1.5B AVG  \textbf{(VD)}  & 57.42 & 45.31 & 57.42 & 49.22\\
        \cmidrule{1-5}
        reranker MAX  \textbf{(VD)}  & - & - & 57.42 & 44.53 \\
        reranker AVG  \textbf{(VD)}   & - & - & \underline{58.59} & 48.05 \\
        \cmidrule{1-5}
        LLM-as-a-Judge MAX  \textbf{(LLM)} & 48.44 & 52.73 & 41.41 & 47.27\\
        LLM-as-a-Judge AVG  \textbf{(LLM)} & 58.20 & \underline{57.81} & 57.03 & \underline{57.68} \\
        \cmidrule{1-5}
        \textbf{PD$^3$ feedback (Ours)} & \textbf{67.97} & \textbf{64.45} & \textbf{68.36} &  \textbf{64.58} \\
        \bottomrule
      \end{tabular}
    }
\end{table}

\vpara{Analysis of Task 2 (RQ2)} Table~\ref{scoring-exp-results} presents the experimental results for Task 2\footnote{Controlled rerankers' scores are omitted because fixed candidate sets are used.}. PD$^3$ achieves the best results under both the Controlled and End-to-End settings, and McNemar's tests in Table~\ref{tab:statistical_test_task2_human} and Table~\ref{tab:statistical_test_task2} confirm that these gains are statistically significant. Under the Controlled setting, where all methods score the same expert-annotated reference sets, PD$^3$ improves ACC from the strongest baseline's 61.33 to 67.97 and Weighted ACC from 57.81 to 64.45. This shows that PD$^3$'s set-level feedback module is better at translating a relevant reference set into a duplication score than max/average aggregation over pairwise similarity scores.

The End-to-End setting further demonstrates that better retrieval improves downstream feedback quality. When each method uses its own retrieved references, PD$^3$ reaches 68.36 ACC and 64.58 Weighted ACC, exceeding the strongest baselines by 9.77 and 6.90 percentage points, respectively. The larger ACC margin in the End-to-End setting than in the Controlled setting suggests that PD$^3$'s retrieval advantage is propagated to the final scoring stage. This result directly supports the pipeline motivation in Section~\ref{section-frameworkdesign-mad}: accurate many-to-many reference set selection is not an isolated retrieval objective, but a prerequisite for reliable duplication feedback.

The Weighted ACC results also reveal a weakness of single-dimension similarity scoring. Several VD baselines obtain reasonable standard ACC but drop sharply under Weighted ACC, such as gte-1.5B MAX decreasing from 61.33 ACC to 41.41 Weighted ACC in the Controlled setting. This suggests that selecting the maximum or average similarity score over top-5 references can overemphasize a single surface-similar project and fail to reflect the multi-dimensional disagreements among expert reviewers. In contrast, PD$^3$ evaluates the project against the whole selected reference set under expert-defined criteria. This approach better aligns with the scenario where experts need to comprehensively compare the project from different perspectives and leads to stronger alignment with human reviewer judgments.


\subsection{Ablation Study}
\label{section-exp-ablation}

\vpara{Ablation Study Settings} To evaluate the effectiveness of each component in PD$^3$ retrieval and comparison fairness of other scheduling variants, we evaluate two groups of variants. The first group weakens the reasoning module: \textit{PD$^3$ w/o MAD} \textbf{(VD)} directly uses preliminary retrieval results, and \textit{PD$^3$ w/o MAD, with voting} \textbf{(LLM Voting)} replaces MAD with one LLM while keeping the same round-robin partition and voting aggregation. The second group keeps MAD but changes the comparison schedule: \textbf{MAD Vanilla} evaluates all 30 candidates in one overlong context; \textbf{Traversal} keeps 5 winners from 10 candidates and adds 5 new candidates each round; \textbf{Random} samples comparison contexts uniformly; and \textbf{Sliding} uses a circular sliding window. For a fair scheduler comparison, Random and Sliding use the same candidate pool size, context size, and number of groups as PD$^3$: $N=30$, $M=20$, and $G=15$.

\begin{figure}[t]
  \centering
  \includegraphics[width=\linewidth]{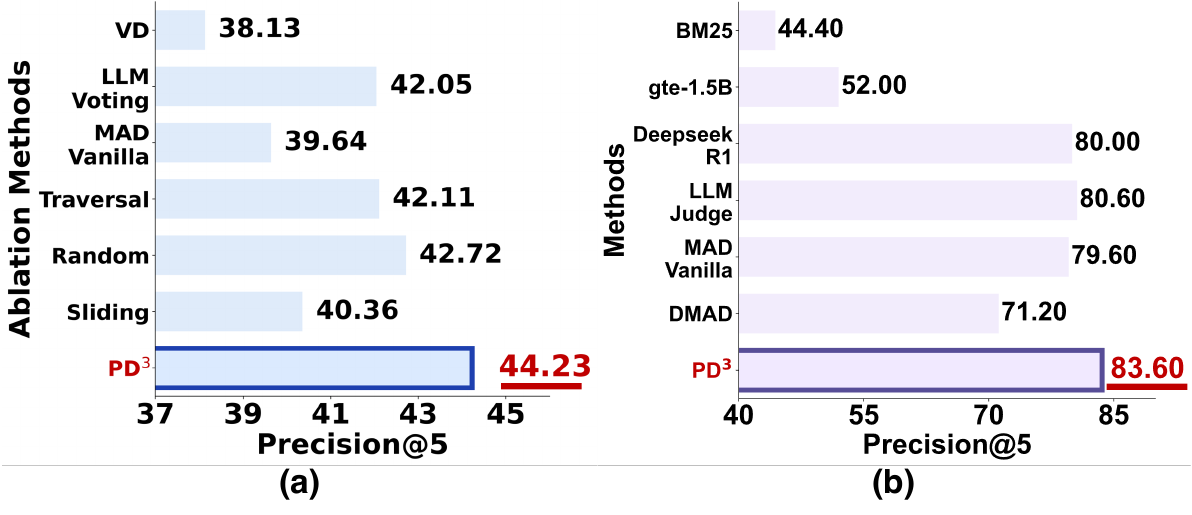}
  \vspace{-0.5cm}
  \caption{Ablation study experiment results (a) and cross-domain experiment results (b).}
  \label{figure-ablation}
  \label{figure-cross-domain}
  \vspace{-0.4cm}
\end{figure}

\begin{figure*}[h]
  \centering
  \includegraphics[width=\textwidth]{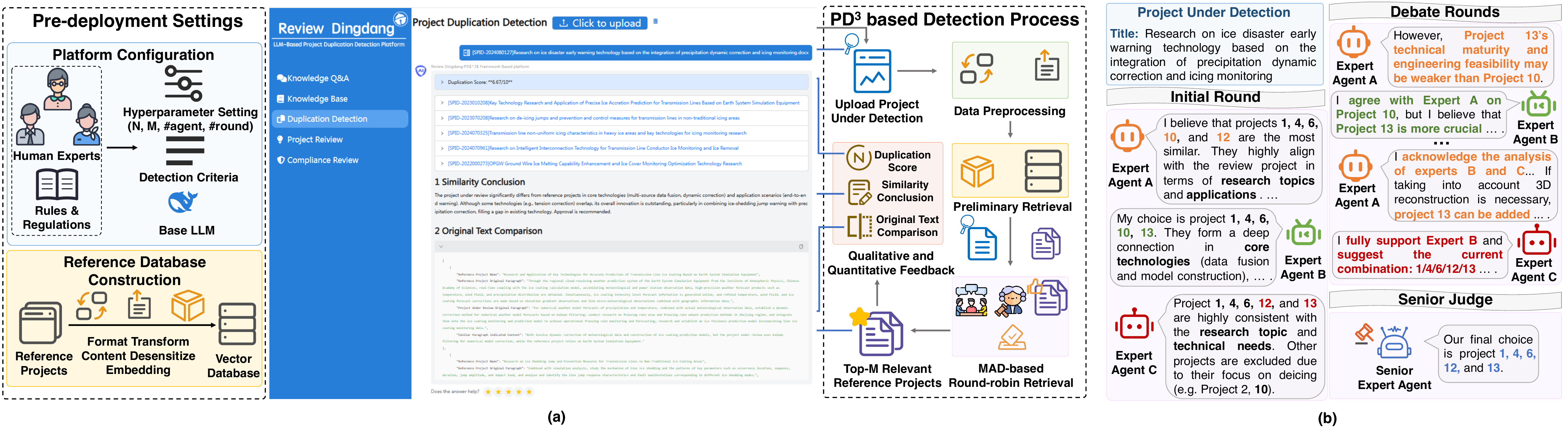}
  \vspace{-0.6cm}
  \caption{(a) \textbf{\textit{Review Dingdang}} platform. Through simple configuration, \textit{Review Dingdang} has been deployed to conduct duplication detection of newly proposed projects and provides feedback to human experts. (b) \textbf{Case study} in round-robin MAD competition. During the debate, Expert A ultimately accepts the views of B and C. A senior judge then makes the final decision.}
  \label{figure-platform}
  \label{figure-case-study}
  \vspace{-0.4cm}
\end{figure*}

\vpara{Ablation Study Analysis (RQ3)} Figure~\ref{figure-ablation} (a) presents the retrieval results. PD$^3$ achieves the best Precision@5, showing that both MAD reasoning and round-robin scheduling are useful. The gap between PD$^3$ and \textbf{VD} confirms the need for LLM-based comparison beyond vector retrieval, while the gap between PD$^3$ and \textbf{LLM Voting} shows that voting alone cannot replace multi-agent debate.

Among scheduling variants, Round-robin gives deterministic fairness guarantees by construction, independent of any hyperparameter setting. For any valid block setting with $J=N/B$ blocks and $r=M/B$ blocks per comparison group, enumerating all $r$-block combinations gives exact exposure balance, $\Delta_{\mathrm{incl}}^{\mathrm{RR}}=0$, a non-zero pairwise lower bound, $\gamma_{\mathrm{pair}}^{\mathrm{RR}}=\binom{J-2}{r-2}$ when $r\ge2$, and bounded context overlap, $o_{\max}^{\mathrm{RR}}=B(r-1)$. We instantiate these expressions with the main setting only to make the comparison concrete: $\gamma_{\mathrm{pair}}^{\mathrm{RR}}=6$ and $o_{\max}^{\mathrm{RR}}=15$. By contrast, Random keeps the same $G=15$ but is only fair in expectation, with $\mathbb{E}[\Delta_{\mathrm{incl}}^{\mathrm{Rand}}]\approx7.39$, $\mathbb{E}[L_{\mathrm{pair}}^{\mathrm{Rand}}(6)]\approx128.60$, and $\mathbb{E}[R_{\mathrm{ctx}}^{\mathrm{Rand}}(15)]\approx4.06$. Sliding also keeps $G=15$, but the unit-step windows create positional bias: $\Delta_{\mathrm{incl}}^{\mathrm{SW}}=15-5=10$, $\gamma_{\mathrm{pair}}^{\mathrm{SW}}=0$, $L_{\mathrm{pair}}^{\mathrm{SW}}(6)=204$, $o_{\max}^{\mathrm{SW}}=19$, and $R_{\mathrm{ctx}}^{\mathrm{SW}}(15)=50$. Traversal is path-dependent, with $\Delta_{\mathrm{incl}}^{\mathrm{Trav}}=1$--$4$ and $\gamma_{\mathrm{pair}}^{\mathrm{Trav}}=0$ in the current setting. 

These results show that PD$^3$ improves retrieval not merely by changing the number of LLM calls, but by providing deterministic schedule-level fairness under the same context budget. Future deployments can increase $N$ only as needed and jointly tune $B$ and $M$ to keep the context size and the number of comparison groups. Random, Sliding, and Traversal do not provide the same deterministic realized-schedule guarantees under a fixed budget. Complete ablation results are in Table~\ref{tab:ablation}. Complete fairness derivations and variant-specific calculations are provided in Appendix~\ref{section-appendix-scheduler-fairness}.

\subsection{Cross-domain Performance}
\label{section-exp-cross-domain}

A critical challenge in evaluating cross-domain project duplication detection is the severe lack of dedicated, open-source datasets. Therefore, we construct a specialized cross-domain benchmark based on bigPatent~\citep{sharma2019bigpatent}, an authentic U.S. patent corpus. To rigorously simulate real-world duplicate patent proposals, we sample 1,000 mechanical patents and use five state-of-the-art LLMs, carefully prompted with expert-level patent-drafting instructions to rewrite 100 of them. This multi-model synthesis mimics sophisticated human plagiarism, ensuring a credible evaluation.

As illustrated in Figure~\ref{figure-cross-domain} (b), PD$^3$ consistently demonstrates superior performance, yielding a 3.00\% improvement over the strongest baseline. This validates the robust cross-domain applicability of PD$^3$. Detailed dataset processing procedure and results are provided in Appendix~\ref{section-appendix-dataset} and Appendix~\ref{section-appendix-expresults}. Additionally, to bridge the gap between synthetic and real-world data, we also conduct experiments on the real-world WikiPassageQA~\citep{cohen2018wikipassageqa} dataset in Appendix~\ref{section-appendix-exp-cross-domain}.

\section{Application}
\label{section-application}

\vpara{Platform and Case Study} Based on PD$^3$, we develop \textit{Review Dingdang},  an online platform for power project duplication detection. As illustrated in Figure~\ref{figure-platform} (a), the platform architecture supports pre-deployment configuration by domain experts through an intuitive interface. Upon submission of a newly proposed project, the platform automatically performs duplication analysis leveraging PD$^3$, subsequently presenting comprehensive detection results, including project under detection, relevant reference projects, and quantitative-qualitative feedback. Notably, \textit{Review Dingdang} incorporates a human-in-the-loop mechanism that allows experts to iteratively refine system performance. This is achieved through two primary intervention modalities: (i) dynamic adjustment of debate prompt rules in the PD$^3$ framework, and (ii) direct denotation feedback on failure cases to optimize the platform's detection algorithms. Figure~\ref{figure-platform} (b) provides a case study in one comparison group and a more detailed analysis is provided in Appendix~\ref{section-case-study}.

\vpara{Application Impact} With \textit{Review Dingdang}, human experts have conducted duplication detection on \textbf{442} newly proposed projects applying for SGCC scientific funding in 2025. The platform helped experts detect \textbf{41} ineligible projects highly duplicate of historical projects (9.28\%), saving 13.44 million USD, demonstrating its effectiveness and potential social impact. Compared with SGCC’s previous human review process, which requires about 1 hour per project, PD$^3$ requires only \textbf{3 minutes} and \textbf{1.34 USD} per project. Particularly, in all live detection tests in 2025, \textit{Review Dingdang} detected 24 more highly-duplicate projects (8.09 million USD) and saved over 400 person-hours. This represents a 9-fold reduction in manual time and roughly a 100× reduction in cost. Runtime performance and cost-effectiveness analysis are provided in Appendix~\ref{section-exp-runningtime}.

\section{Conclusion}
\label{section-conclusion}

We present PD$^3$, a framework for project duplication detection via adapted multi-agent debate. We reframe key task as many-to-many set selection rather than strict similarity ranking. PD$^3$ adapts MAD to long technical proposals by combining local expert-agent deliberation with global round-robin scheduling, so candidates receive balanced exposure while each debate stays within a manageable context. Furthermore, PD$^3$ provides both quantitative and qualitative feedback. Ultimately, validation using real-world data and online deployment confirms its utility and also illustrates a broader perspective: addressing complex real-world problems hinges on formalizing key trade-offs to unlock practical solutions.




\bibliographystyle{ACM-Reference-Format}
\bibliography{ref}

@manual{CN108829780B,
author = {Yu, Yang and Liu, Lei and Xu, Xiangyi and Bai, Shaoqian},
title = {Text detection methods, devices, computing equipment, and computer-readable storage media},
edition = {CN108829780B},
year = {2022},
pages = {18}
}

@manual{patent_2,
author = {Zhang, Zhenhai and Sun, Xiongyong},
title = {A method and system for automatically detecting academic misconduct literature},
edition = {CN101833579B},
year = {2012},
pages = {10}
}

@article{liu2023lost,
  title={Lost in the middle: How language models use long contexts},
  author={Liu, Nelson F and Lin, Kevin and Hewitt, John and Paranjape, Ashwin and Bevilacqua, Michele and Petroni, Fabio and Liang, Percy},
  journal={arXiv preprint arXiv:2307.03172},
  year={2023}
}

@inproceedings{du2023improving,
  title={Improving factuality and reasoning in language models through multiagent debate},
  author={Du, Yilun and Li, Shuang and Torralba, Antonio and Tenenbaum, Joshua B and Mordatch, Igor},
  booktitle={Forty-first International Conference on Machine Learning},
  year={2023}
}

@inproceedings{bensalem2014intrinsic,
  title={Intrinsic plagiarism detection using n-gram classes},
  author={Bensalem, Imene and Rosso, Paolo and Chikhi, Salim},
  booktitle={Proceedings of the 2014 Conference on Empirical Methods in Natural Language Processing},
  pages={1459--1464},
  year={2014}
}

@inproceedings{liang2023encouraging,
  title={Encouraging Divergent Thinking in Large Language Models through Multi-Agent Debate},
  author={Liang, Tian and He, Zhiwei and Jiao, Wenxiang and Wang, Xing and Wang, Yan and Wang, Rui and Yang, Yujiu and Shi, Shuming and Tu, Zhaopeng},
  booktitle={Proceedings of the 2024 Conference on Empirical Methods in Natural Language Processing},
  pages={17889--17904},
  year={2024}
}

@inproceedings{wang2024debate,
  title={Debate as Optimization: Adaptive Conformal Prediction and Diverse Retrieval for Event Extraction},
  author={Wang, Sijia and Huang, Lifu},
  booktitle={Proceedings of the conference Association for Computational Linguistics Meeting},
  volume={1},
  number={2024},
  year={2024},
  organization={Association for Computational Linguistics}
}

@article{chan2023chateval,
  title={Chateval: Towards better llm-based evaluators through multi-agent debate},
  author={Chan, Chi-Min and Chen, Weize and Su, Yusheng and Yu, Jianxuan and Xue, Wei and Zhang, Shanghang and Fu, Jie and Liu, Zhiyuan},
  journal={arXiv preprint arXiv:2308.07201},
  year={2023}
}

@article{liu2024groupdebate,
  title={Groupdebate: Enhancing the efficiency of multi-agent debate using group discussion},
  author={Liu, Tongxuan and Wang, Xingyu and Huang, Weizhe and Xu, Wenjiang and Zeng, Yuting and Jiang, Lei and Yang, Hailong and Li, Jing},
  journal={arXiv preprint arXiv:2409.14051},
  year={2024}
}

@inproceedings{pham2023let,
  title={Let Models Speak Ciphers: Multiagent Debate through Embeddings},
  author={Pham, Chau and Liu, Boyi and Yang, Yingxiang and Chen, Zhengyu and Liu, Tianyi and Yuan, Jianbo and Plummer, Bryan A and Wang, Zhaoran and Yang, Hongxia},
  year={2024},
  booktitle={The Twelfth International Conference on Learning Representations}
}

@article{gottweis2025towards,
  title={Towards an AI co-scientist},
  author={Gottweis, Juraj and Weng, Wei-Hung and Daryin, Alexander and Tu, Tao and Palepu, Anil and Sirkovic, Petar and Myaskovsky, Artiom and Weissenberger, Felix and Rong, Keran and Tanno, Ryutaro and {others}},
  journal={arXiv preprint arXiv:2502.18864},
  year={2025}
}

@inproceedings{chen2025debate,
  title={Debate-Feedback: A Multi-Agent Framework for Efficient Legal Judgment Prediction},
  author={Chen, Xi and Mao, Mao and Li, Shuo and Shangguan, Haotian},
  booktitle={Proceedings of the 2025 Conference of the Nations of the Americas Chapter of the Association for Computational Linguistics: Human Language Technologies (Volume 2: Short Papers)},
  pages={462--470},
  year={2025}
}

@article{estornell2024multi,
  title={Multi-LLM debate: Framework, principals, and interventions},
  author={Estornell, Andrew and Liu, Yang},
  journal={Advances in Neural Information Processing Systems},
  volume={37},
  pages={28938--28964},
  year={2024}
}

@article{kaesberg2025voting,
  title={Voting or Consensus? Decision-Making in Multi-Agent Debate},
  author={Kaesberg, Lars Benedikt and Becker, Jonas and Wahle, Jan Philip and Ruas, Terry and Gipp, Bela},
  journal={arXiv preprint arXiv:2502.19130},
  year={2025}
}

@article{gu2024survey,
  title={A survey on llm-as-a-judge},
  author={Gu, Jiawei and Jiang, Xuhui and Shi, Zhichao and Tan, Hexiang and Zhai, Xuehao and Xu, Chengjin and Li, Wei and Shen, Yinghan and Ma, Shengjie and Liu, Honghao and {others}},
  journal={arXiv preprint arXiv:2411.15594},
  year={2024}
}

@article{zhang2025re2consistencyensureddatasetfullstage,
  title={Re$^2$: A Consistency-ensured Dataset for Full-stage Peer Review and Multi-turn Rebuttal Discussions}, 
  author={Zhang, Daoze and Bao, Zhijian and Du, Sihang and Zhao, Zhiyi and Zhang, Kuangling and Bao, Dezheng and Yang, Yang},
  journal={arXiv preprint arXiv:2505.07920},
  year={2025},
  eprint={}
}

@misc{sgcc2023report,
  title = {The State Grid Corporation of China 2023 Social Responsibility Report.},
  author = {{The State Grid Corporation of China}},
  year         = {2024},
  url = {http://www.sgcc.com.cn/u/cms/sgcc_main/other/202602/d08c469cd7ff47fbab82efbb9c67e10a.pdf},
  note         = {Accessed: 2026-02-18}
}

@misc{doesmartgrid,
  title = {Smart Grid Grants | Department of Energy},
  author = {{The U.S. Department of Energy}},
  year         = {2023},
  url = {https://www.energy.gov/gdo/smart-grid-grants},
  note         = {Accessed: 2026-02-18}
}

@inproceedings{shen2021mred,
  title={MReD: A Meta-Review Dataset for Structure-Controllable Text Generation},
  author={Shen, Chenhui and Cheng, Liying and Zhou, Ran and Bing, Lidong and You, Yang and Si, Luo},
  booktitle={Findings of the Association for Computational Linguistics: ACL 2022},
  pages={2521--2535},
  year={2022}
}

@inproceedings{smit2024should,
  title={Should we be going MAD? a look at multi-agent debate strategies for LLMs},
  author={Smit, Andries and Grinsztajn, Nathan and Duckworth, Paul and Barrett, Thomas D and Pretorius, Arnu},
  booktitle={Proceedings of the 41st International Conference on Machine Learning},
  pages={45883--45905},
  year={2024}
}

@misc{volcengine,
  title        = {volcengine},
  author       = {ByteDance},
  year         = {2025},
  url          = {https://www.volcengine.com/},
  note         = {Accessed: 2026-02-18}
}

@inproceedings{lin2004rouge,
  title={Rouge: A package for automatic evaluation of summaries},
  author={Lin, Chin-Yew},
  booktitle={Text summarization branches out},
  pages={74--81},
  year={2004}
}

@article{robertson2009probabilistic,
  title={The probabilistic relevance framework: BM25 and beyond},
  author={Robertson, Stephen and Zaragoza, Hugo and {others}},
  journal={Foundations and Trends{\textregistered} in Information Retrieval},
  volume={3},
  number={4},
  pages={333--389},
  year={2009},
  publisher={Now Publishers, Inc.}
}

@article{li2023making,
  title={Making large language models a better foundation for dense retrieval},
  author={Li, Chaofan and Liu, Zheng and Xiao, Shitao and Shao, Yingxia},
  journal={arXiv preprint arXiv:2312.15503},
  year={2023}
}

@article{chen2024bge,
  title={Bge m3-embedding: Multi-lingual, multi-functionality, multi-granularity text embeddings through self-knowledge distillation},
  author={Chen, Jianlv and Xiao, Shitao and Zhang, Peitian and Luo, Kun and Lian, Defu and Liu, Zheng},
  journal={arXiv preprint arXiv:2402.03216},
  year={2024}
}

@article{zheng2023judging,
  title={Judging llm-as-a-judge with mt-bench and chatbot arena},
  author={Zheng, Lianmin and Chiang, Wei-Lin and Sheng, Ying and Zhuang, Siyuan and Wu, Zhanghao and Zhuang, Yonghao and Lin, Zi and Li, Zhuohan and Li, Dacheng and Xing, Eric and {others}},
  journal={Advances in Neural Information Processing Systems},
  volume={36},
  pages={46595--46623},
  year={2023}
}

@article{guo2025deepseek,
  title={Deepseek-r1: Incentivizing reasoning capability in llms via reinforcement learning},
  author={Guo, Daya and Yang, Dejian and Zhang, Haowei and Song, Junxiao and Zhang, Ruoyu and Xu, Runxin and Zhu, Qihao and Ma, Shirong and Wang, Peiyi and Bi, Xiao and {others}},
  journal={arXiv preprint arXiv:2501.12948},
  year={2025}
}

@inproceedings{wang2024mmlu,
  title={Mmlu-pro: A more robust and challenging multi-task language understanding benchmark},
  author={Wang, Yubo and Ma, Xueguang and Zhang, Ge and Ni, Yuansheng and Chandra, Abhranil and Guo, Shiguang and Ren, Weiming and Arulraj, Aaran and He, Xuan and Jiang, Ziyan and {others}},
  booktitle={The Thirty-eight Conference on Neural Information Processing Systems Datasets and Benchmarks Track},
  year={2024}
}

@article{deepseekai2024deepseekv3technicalreport,
  title={Deepseek-v3 technical report},
  author={Liu, Aixin and Feng, Bei and Xue, Bing and Wang, Bingxuan and Wu, Bochao and Lu, Chengda and Zhao, Chenggang and Deng, Chengqi and Zhang, Chenyu and Ruan, Chong and {others}},
  journal={arXiv preprint arXiv:2412.19437},
  year={2024}
}

@article{sturua2024jina,
  title={jina-embeddings-v3: Multilingual embeddings with task lora},
  author={Sturua, Saba and Mohr, Isabelle and Akram, Mohammad Kalim and G{\"u}nther, Michael and Wang, Bo and Krimmel, Markus and Wang, Feng and Mastrapas, Georgios and Koukounas, Andreas and Wang, Nan and {others}},
  journal={arXiv preprint arXiv:2409.10173},
  year={2024}
}

@article{mikolov2013efficient,
  title={Efficient estimation of word representations in vector space},
  author={Mikolov, Tomas and Chen, Kai and Corrado, Greg and Dean, Jeffrey},
  journal={arXiv preprint arXiv:1301.3781},
  year={2013}
}

@inproceedings{xiong2023examining,
  title={Examining Inter-Consistency of Large Language Models Collaboration: An In-depth Analysis via Debate},
  author={Xiong, Kai and Ding, Xiao and Cao, Yixin and Liu, Ting and Qin, Bing},
  booktitle={The 2023 Conference on Empirical Methods in Natural Language Processing},
  year={2023}
}

@inproceedings{liu2025breaking,
  title={Breaking mental set to improve reasoning through diverse multi-agent debate},
  author={Liu, Yexiang and Cao, Jie and Li, Zekun and He, Ran and Tan, Tieniu},
  booktitle={The Thirteenth International Conference on Learning Representations},
  year={2025}
}

@article{qwen3embedding,
  title={Qwen3 Embedding: Advancing Text Embedding and Reranking Through Foundation Models},
  author={Zhang, Yanzhao and Li, Mingxin and Long, Dingkun and Zhang, Xin and Lin, Huan and Yang, Baosong and Xie, Pengjun and Yang, An and Liu, Dayiheng and Lin, Junyang and Huang, Fei and Zhou, Jingren},
  journal={arXiv preprint arXiv:2506.05176},
  year={2025}
}

@article{wei2022chain,
  title={Chain-of-thought prompting elicits reasoning in large language models},
  author={Wei, Jason and Wang, Xuezhi and Schuurmans, Dale and Bosma, Maarten and Xia, Fei and Chi, Ed and Le, Quoc V and Zhou, Denny and {others}},
  journal={Advances in neural information processing systems},
  volume={35},
  pages={24824--24837},
  year={2022}
}

@article{zheng2023take,
  title={Take a step back: Evoking reasoning via abstraction in large language models},
  author={Zheng, Huaixiu Steven and Mishra, Swaroop and Chen, Xinyun and Cheng, Heng-Tze and Chi, Ed H and Le, Quoc V and Zhou, Denny},
  journal={arXiv preprint arXiv:2310.06117},
  year={2023}
}

@article{madaan2023self,
  title={Self-refine: Iterative refinement with self-feedback},
  author={Madaan, Aman and Tandon, Niket and Gupta, Prakhar and Hallinan, Skyler and Gao, Luyu and Wiegreffe, Sarah and Alon, Uri and Dziri, Nouha and Prabhumoye, Shrimai and Yang, Yiming and {others}},
  journal={Advances in Neural Information Processing Systems},
  volume={36},
  pages={46534--46594},
  year={2023}
}

@article{kim2023language,
  title={Language models can solve computer tasks},
  author={Kim, Geunwoo and Baldi, Pierre and McAleer, Stephen},
  journal={Advances in Neural Information Processing Systems},
  volume={36},
  pages={39648--39677},
  year={2023}
}

@inproceedings{kim2024debate,
  title={DEBATE: Devil’s Advocate-Based Assessment and Text Evaluation},
  author={Kim, Alex and Kim, Keonwoo and Yoon, Sangwon},
  booktitle={Findings of the Association for Computational Linguistics ACL 2024},
  pages={1885--1897},
  year={2024}
}

@inproceedings{chen2024reconcile,
  title={ReConcile: Round-Table Conference Improves Reasoning via Consensus among Diverse LLMs},
  author={Chen, Justin and Saha, Swarnadeep and Bansal, Mohit},
  booktitle={Proceedings of the 62nd Annual Meeting of the Association for Computational Linguistics (Volume 1: Long Papers)},
  pages={7066--7085},
  year={2024}
}

@inproceedings{yin2023exchange,
  title={Exchange-of-Thought: Enhancing Large Language Model Capabilities through Cross-Model Communication},
  author={Yin, Zhangyue and Sun, Qiushi and Chang, Cheng and Guo, Qipeng and Dai, Junqi and Huang, Xuan-Jing and Qiu, Xipeng},
  booktitle={Proceedings of the 2023 Conference on Empirical Methods in Natural Language Processing},
  pages={15135--15153},
  year={2023}
}

@inproceedings{chen2025tourrank,
  title={Tourrank: Utilizing large language models for documents ranking with a tournament-inspired strategy},
  author={Chen, Yiqun and Liu, Qi and Zhang, Yi and Sun, Weiwei and Ma, Xinyu and Yang, Wei and Shi, Daiting and Mao, Jiaxin and Yin, Dawei},
  booktitle={Proceedings of the ACM on Web Conference 2025},
  pages={1638--1652},
  year={2025}
}

@inproceedings{sharma2019bigpatent,
  title={BIGPATENT: A Large-Scale Dataset for Abstractive and Coherent Summarization},
  author={Sharma, Eva and Li, Chen and Wang, Lu},
  booktitle={Proceedings of the 57th Annual Meeting of the Association for Computational Linguistics},
  pages={2204--2213},
  year={2019}
}

@article{zeng2025glm,
  title={Glm-4.5: Agentic, reasoning, and coding (arc) foundation models},
  author={Zeng, Aohan and Lv, Xin and Zheng, Qinkai and Hou, Zhenyu and Chen, Bin and Xie, Chengxing and Wang, Cunxiang and Yin, Da and Zeng, Hao and Zhang, Jiajie and {others}},
  journal={arXiv preprint arXiv:2508.06471},
  year={2025}
}

@article{qwen3technicalreport,
  title={Qwen3 technical report},
  author={Yang, An and Li, Anfeng and Yang, Baosong and Zhang, Beichen and Hui, Binyuan and Zheng, Bo and Yu, Bowen and Gao, Chang and Huang, Chengen and Lv, Chenxu and {others}},
  journal={arXiv preprint arXiv:2505.09388},
  year={2025}
}

@article{qwen2.5-1m,
      title={Qwen2.5-1M Technical Report}, 
      author={An Yang and Bowen Yu and Chengyuan Li and Dayiheng Liu and Fei Huang and Haoyan Huang and Jiandong Jiang and Jianhong Tu and Jianwei Zhang and Jingren Zhou and Junyang Lin and Kai Dang and Kexin Yang and Le Yu and Mei Li and Minmin Sun and Qin Zhu and Rui Men and Tao He and Weijia Xu and Wenbiao Yin and Wenyuan Yu and Xiafei Qiu and Xingzhang Ren and Xinlong Yang and Yong Li and Zhiying Xu and Zipeng Zhang},
      journal={arXiv preprint arXiv:2501.15383},
      year={2025}
}

@misc{gpt5openai,
  title        = {Introducing GPT-5},
  author       = {OpenAI},
  year         = {2025},
  url          = {https://openai.com/index/introducing-gpt-5/},
  note         = {Accessed: 2026-02-18}
}

@misc{gemini3prodeepmind,
  title        = {Gemini 3 Pro},
  author       = {Google Deepmind},
  year         = {2025},
  url          = {https://deepmind.google/models/gemini/pro/},
  note         = {Accessed: 2026-02-18}
}

@misc{claudesonnet45deepmind,
  title = {Introducing Claude Sonnet 4.5},
  author = {Anthropic},
  year         = {2025},
  url = {https://www.anthropic.com/news/claude-sonnet-4-5},
  note         = {Accessed: 2026-02-18}
}

@article{fan2025imad,
  title={{iMAD}: Intelligent Multi-Agent Debate for Efficient and Accurate LLM Inference},
  author={Fan, Wei and Yoon, JinYi and Ji, Bo},
  journal={arXiv preprint arXiv:2511.11306},
  year={2025}
}

@article{wynn2025talk,
  title={Talk Isn't Always Cheap: Understanding Failure Modes in Multi-Agent Debate},
  author={Wynn, Andrea and Satija, Harsh and Hadfield, Gillian},
  journal={arXiv preprint arXiv:2509.05396},
  year={2025}
}

@article{yang2025revisiting,
  title={Revisiting Multi-Agent Debate as Test-Time Scaling: A Systematic Study of Conditional Effectiveness},
  author={Yang, Yongjin and Yi, Euiin and Ko, Jongwoo and Lee, Kimin and Jin, Zhijing and Yun, Se-Young},
  journal={arXiv preprint arXiv:2505.22960},
  year={2025}
}

@inproceedings{oh2025debate,
  title={When Debate Fails: Bias Reinforcement in Large Language Models},
  author={Oh, Jihwan and Jeong, Minchan and Ko, Jongwoo and Yun, Se-Young},
  booktitle={Workshop on Reasoning and Planning for Large Language Models},
  year={2025}
}

@inproceedings{cohen2018wikipassageqa,
  title={Wikipassageqa: A benchmark collection for research on non-factoid answer passage retrieval},
  author={Cohen, Daniel and Yang, Liu and Croft, W Bruce},
  booktitle={The 41st international ACM SIGIR conference on research \& development in information retrieval},
  pages={1165--1168},
  year={2018}
}

@article{neelakantan2022text,
  title={Text and code embeddings by contrastive pre-training},
  author={Neelakantan, Arvind and Xu, Tao and Puri, Raul and Radford, Alec and Han, Jesse Michael and Tworek, Jerry and Yuan, Qiming and Tezak, Nikolas and Kim, Jong Wook and Hallacy, Chris and others},
  journal={arXiv preprint arXiv:2201.10005},
  year={2022}
}

@article{tseng2020effective,
  title={Effective FAQ retrieval and question matching with unsupervised knowledge injection},
  author={Tseng, Wen-Ting and Lo, Tien-Hong and Hsu, Yung-Chang and Chen, Berlin},
  journal={arXiv preprint arXiv:2010.14049},
  year={2020}
}

@inproceedings{
    hua2026context,
    title={Context Learning for Multi-Agent Discussion},
    author={Xingyuan Hua and Sheng Yue and Xinyi Li and Yizhe Zhao and Jinrui Zhang and Ju Ren},
    booktitle={The Fourteenth International Conference on Learning Representations},
    year={2026},
    url={https://openreview.net/forum?id=EUu8TILWpR}
}

@article{liu2026prepare,
  title={Prepare Reasoning Language Models for Multi-Agent Debate with Self-Debate Reinforcement Learning},
  author={Liu, Chenxi and Chen, Yanshuo and Chen, Ruibo and Xiong, Tianyi and Zheng, Tong and Huang, Heng},
  journal={arXiv preprint arXiv:2601.22297},
  year={2026}
}

@inproceedings{xiao2022oagbert,
author = {Liu, Xiao and Yin, Da and Zheng, Jingnan and Zhang, Xingjian and Zhang, Peng and Yang, Hongxia and Dong, Yuxiao and Tang, Jie},
title = {OAG-BERT: Towards a Unified Backbone Language Model for Academic Knowledge Services},
year = {2022},
isbn = {9781450393850},
publisher = {Association for Computing Machinery},
address = {New York, NY, USA},
url = {https://doi.org/10.1145/3534678.3539210},
doi = {10.1145/3534678.3539210},
booktitle = {Proceedings of the 28th ACM SIGKDD Conference on Knowledge Discovery and Data Mining},
pages = {3418–3428},
numpages = {11},
location = {Washington DC, USA},
series = {KDD '22}
}

@inproceedings{cohan-etal-2020-specter,
    title = "{SPECTER}: Document-level Representation Learning using Citation-informed Transformers",
    author = "Cohan, Arman  and
      Feldman, Sergey  and
      Beltagy, Iz  and
      Downey, Doug  and
      Weld, Daniel",
    editor = "Jurafsky, Dan  and
      Chai, Joyce  and
      Schluter, Natalie  and
      Tetreault, Joel",
    booktitle = "Proceedings of the 58th Annual Meeting of the Association for Computational Linguistics",
    month = jul,
    year = "2020",
    address = "Online",
    publisher = "Association for Computational Linguistics",
    url = "https://aclanthology.org/2020.acl-main.207/",
    doi = "10.18653/v1/2020.acl-main.207",
    pages = "2270--2282"
}

\appendix

\section{Related work}
\label{section-appendix-relatedwork}

\subsection{Textual Duplication Detection}
\label{section-appendix-relatedwork-duplicationdetection}
Textual duplication detection is a crucial computational linguistics task for identifying content replication across documents, playing a vital role in protecting academic integrity and intellectual property. Early methods primarily relied on lexical-level analysis, such as n-gram overlap quantification~\citep{bensalem2014intrinsic} and dynamic programming algorithms like Smith-Waterman for local alignment. Notably, ~\citet{CN108829780B} introduced advanced syntactic-level processing—including sentence segmentation, lexical decomposition, and TF-IDF differential computation—for similarity matrix construction, as implemented in the Wanfang duplication detection platform. 

While these character-matching methods achieve high precision in verbatim detection, they lack high-level semantic comprehension, making them susceptible to evasion through paraphrasing (e.g., synonym substitution and syntactic restructuring). Later advancements adopt distributed semantic representations, employing topic modeling (e.g., Latent Dirichlet Allocation) and word embeddings (e.g., Word2Vec~\citep{mikolov2013efficient}) for document similarity. ~\citet{patent_2} proposed a hierarchical detection framework combining document-level keyword ranking with sentence-level synonymy detection, widely used in the CNKI duplication detection platform. Although these approaches capture surface-level semantic relationships, they struggle with complex semantic transformations. The rise of large language models has revolutionized reasoning tasks~\citep{gu2024survey}, yet their mechanisms inherently conflict with the reality that strict ranking task does not match the detection requirements.

\subsection{Multi-Agent Debate Systems}
\label{section-appendix-relatedwork-mad}
Multi-Agent Debate (MAD)~\citep{liang2023encouraging, du2023improving, xiong2023examining, chen2024reconcile} implements the ``society of minds'' framework through collaborative interactions among LLM agents. This approach addresses key limitations of single-model reasoning—such as confirmation bias, hallucinations, and logical inconsistencies—through iterative adversarial knowledge refinement. Empirical studies show that MAD improves reasoning via three mechanisms: (1) collective error correction, (2) perspective diversification, and (3) systematic reasoning reinforcement.

MAD has proven effective in diverse applications. In scientific discovery, ~\citet{gottweis2025towards} use tournament-style debates to generate and refine biomedical hypotheses. For legal judgment prediction, ~\citet{chen2025debate} combine MAD with reliability assessment to reduce reliance on large datasets. In event extraction, the Debate as Optimization (DAO) system ~\citep{wang2024debate} iteratively improves outputs without parameter tuning. However, MAD remains unexplored for duplication detection, a gap that our work bridges.

Recent advancements in MAD have primarily focused on four key dimensions: 

(1) \textbf{Communication optimization}, where ~\citet{pham2023let} demonstrate the superiority of embedding-based interaction over natural language debate and ~\citet{yin2023exchange} further improve performance through refined optimization of communication mechanisms. ~\citet{oh2025debate} propose optimizing each debater's speech one-on-one to prevent error propagation. Recent context-learning work further attributes discussion inconsistency to misaligned agent contexts and learns per-agent context generators to organize and refine exchanged information, helping agents avoid premature convergence on majority noise~\citep{hua2026context};

(2) \textbf{Role specialization}, with~\citet{chan2023chateval} and ~\citet{kim2024debate} establishing that heterogeneous agent personas significantly outperform homogeneous configurations. Besides, ~\citet{liu2025breaking} achieve state-of-the-art MAD performance by employing diverse reasoning methods for each participating agent. Several studies show the important improvement of MAD through enhancing the heterogeneity of base models and model capabilities~\citep{yang2025revisiting, wynn2025talk};

(3) \textbf{Decision-making efficiency}, where ~\citet{liu2024groupdebate} introduce grouped debates to reduce computational overhead, while ~\citet{kaesberg2025voting} systematically evaluate voting versus consensus protocols across task types. ~\citet{fan2025imad} adaptively decide whether to initiate debate to reduce token consumption. 

(4) \textbf{Debate-oriented model preparation}, where recent self-debate reinforcement learning explicitly trains reasoning models to act both as standalone solvers and as debate participants. By jointly optimizing initial responses and debate-conditioned revisions with verifiable rewards, this line of work improves MAD performance while also strengthening single-agent reasoning~\citep{liu2026prepare}.

However, these methodological refinements have predominantly targeted question answering scenarios, leaving their applicability to complex, multi-faceted tasks like project duplication detection largely unexplored. Existing MAD lacks schedule-level fairness under context constraints, which motivates our adapted round-robin MAD retrieval.


\section{Preliminary Retrieval Coverage}
\label{section-appendix-preliminary-retrieval}

\begin{figure}[h]
  \centering
  \includegraphics[width=0.75\linewidth]{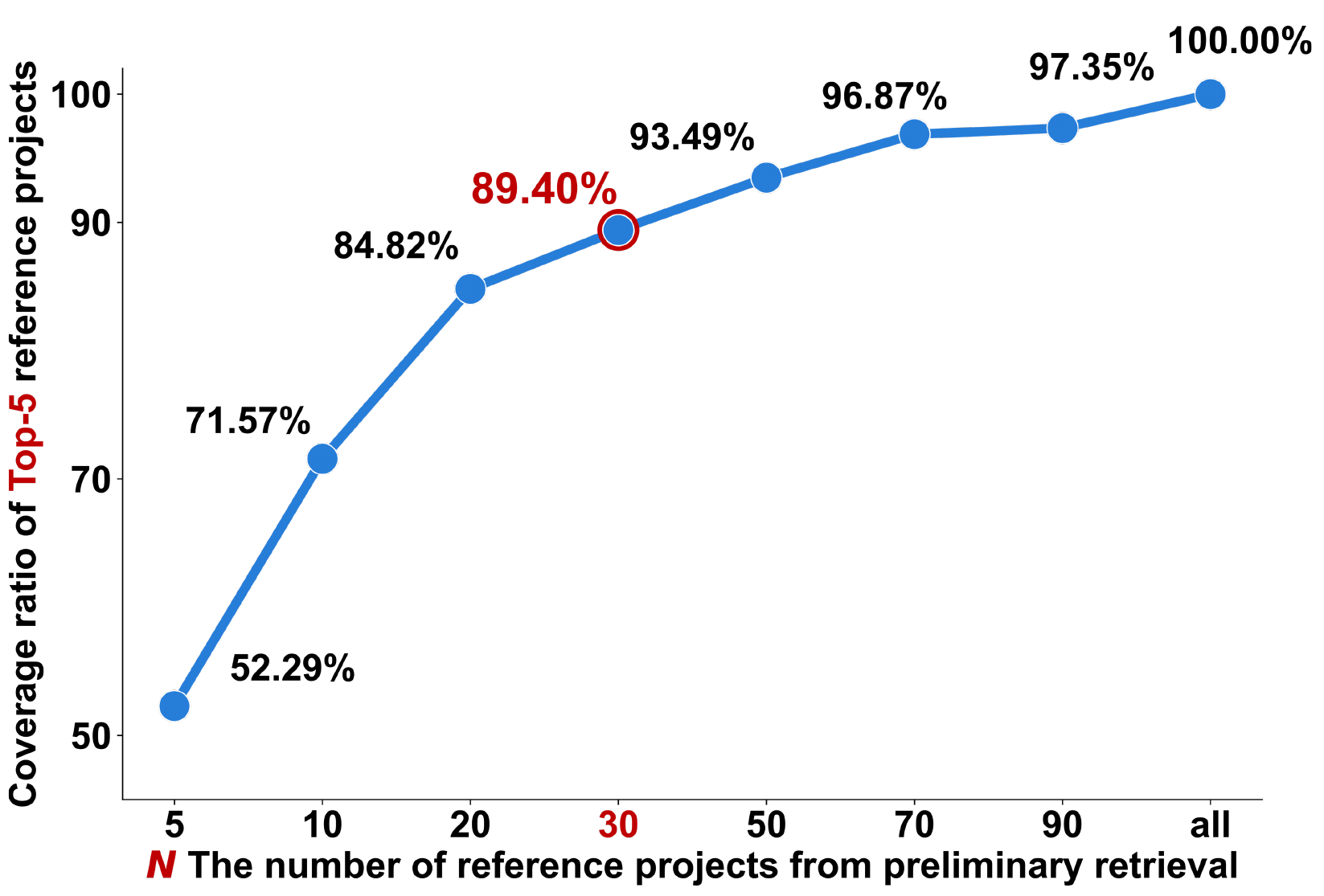}
  \caption{Top-5 reference projects coverage ratio at varying preliminary retrieval numbers $N$.}
  \label{figure-topk}
  \vspace{-0.3cm}
\end{figure}

\vpara{Preliminary Retrieval Coverage Analysis} Figure~\ref{figure-topk} shows how often the preliminary retrieved candidate pool contains human-annotated top-5 reference projects as $N$ increases. The coverage ratio rises sharply from 52.29\% at $N=5$ to 71.57\% at $N=10$ and 84.82\% at $N=20$, indicating that a very small candidate pool would miss many relevant historical projects before LLM-based set selection. Increasing $N$ to 30 further raises coverage to 89.40\%, which is close to the 90\% level needed for high-recall downstream detection. Larger pools provide additional recall, reaching 93.49\%, 96.87\%, and 97.35\% at $N=50$, $70$, and $90$, respectively, and using all candidates reaches 100\%. However, these gains come with substantially larger LLM comparison contexts or more comparison groups. Therefore, we choose $N=30$ as a budget-conscious operating point that preserves most annotated references while keeping the subsequent round-robin MAD retrieval tractable.

\section{Dataset details}
\label{section-appendix-dataset}

\begin{table}[h]
  \caption{Dataset details of power scientific projects.}
  \label{dataset-info}
  \centering
  \resizebox{1\linewidth}{!}{
      \begin{tabular}{crrr}
        \toprule
        \multirow{2}{*}{Year} & Number of  & Average length & Average length \\
         & projects & (in words) & (in tokens)\\
        \midrule
        2022 & 223 & 24,610 & 13,395 \\
        \cmidrule{1-4}
        2023 & 292 & 26,305 & 14,142\\
        \cmidrule{1-4}
        2024 & 318 & 29,680 & 15,851 \\
        \cmidrule{1-4}
        \textbf{Total} & \textbf{833} & \textbf{27,140} & \textbf{14,594}\\
        \bottomrule
      \end{tabular}
    }
\end{table}

\vpara{Power Project Data} The power project data used in this work consists of real projects from the State Grid Corporation of China. We anonymize sensitive information (e.g., applicant details) and randomly generate IDs, retaining only titles and text content. The original documents of these projects are provided in the form of feasibility study reports, often existing in formats such as doc, docx and pdf. The collection, use, and processing of these data have been reviewed by ethics experts and approved for this study. More information is in Table~\ref{dataset-info}. 

\vpara{Cross-domain Data} The patent data used in cross-domain experiments are from F-kind (Mechanical Engineering; Lightning; Heating; Weapons; Blasting) patents in bigPatent~\citep{sharma2019bigpatent}. We randomly sample 1,000 patents with their abstracts. Additionally, we rewrite 100 sampled patents as test cases with 5 different LLMs (GLM-4.6~\citep{zeng2025glm}, Qwen3-235B-A22B-Instruct-2507~\citep{qwen3technicalreport,qwen2.5-1m}, GPT-5~\citep{gpt5openai}, Gemini 3 pro~\citep{gemini3prodeepmind} and Claude Sonnet 4.5~\citep{claudesonnet45deepmind}) to simulate the duplication detection scenarios commonly encountered in the real world. The cross-domain data are publicly available in \href{https://anonymous.4open.science/r/PD-3-KDD}{our repository}.

\vpara{Human Annotation} We invite power-domain experts for power data annotation in Task 1 and Task 2 and compensate them based on the average standard hourly wage according to working hours.

The annotation instructions provided for power-domain experts are: Annotation data consists of a batch of scientific and technological research projects in the field of electricity. Each project involves scientific and technological research related to the power domain. The data provides the project ID, name, and application content. The annotation task is based on human evaluation, selecting the top-5 most relevant projects from the preliminary shortlisted 30-project candidate pool. Detection Criteria: There is duplication in research content, key technologies, or applications. Note that the project data are protected by confidentiality agreements and must not be disseminated.

\section{Cross-domain Experiments Analysis}
\label{section-appendix-exp-cross-domain}

\begin{table}[h]
  \vspace{-0.2cm}
  \caption{Cross-domain experiment results on WikiPassageQA dataset.}
  \vspace{-0.3cm}
  \label{tab:cross-domain_Wiki}
  \centering
  \resizebox{\linewidth}{!}{
      \begin{tabular}{crrr}
        \toprule
        \multirow{2}{*}{\textbf{Method}} & \multirow{2}{*}{\textbf{Recall@5}} & \multicolumn{2}{c}{\textbf{T Test}} \\ 
        & & \textbf{T} & \textbf{P} \\
        \midrule
        ROUGE-L \textbf{(WF)} & 40.59 & 19.7921 & 0.0000  \\
        BM25 \textbf{(WF)} & 67.62 & 10.8516 & 0.0000\\
        \cmidrule{1-4}
        gte-1.5B \textbf{(VD)}  & 83.15 & 5.7394 & 0.0000 \\
        Qwen3-8B as reranker  \textbf{(VD)}  & 84.62 & 5.1018 & 0.0000 \\
        \cmidrule{1-4}
        DeepSeek V3 \textbf{(LLM)} & \underline{90.86} & 2.6027 & 0.0097\\
        DeepSeek R1 \textbf{(LLM)} & 90.49 & 2.4613 & 0.0144\\
        TourRank \textbf{(LLM)} & 90.05 & 3.4669 & 0.0060\\
        \cmidrule{1-4}
        MAD Vanilla \textbf{(MAD)} & 89.49 & 3.6205 & 0.0003\\
        DMAD \textbf{(MAD)} & 87.50 & 4.5526 & 0.0000\\
        \cmidrule{1-4}
        \textbf{PD$^3$ MAD Round-robin (Ours)} & \textbf{92.87} & - & - \\
        \bottomrule
      \end{tabular}
    }
    \vspace{-0.3cm}
\end{table}

We present representative results on 286 cases (all available test instances with more than five candidates and context lengths suitable for model processing), including the best-performing baselines from each category listed in Table~\ref{retrieval-exp-results}. Following the original settings in~\citet{cohen2018wikipassageqa}, we adopt Recall@5 as the evaluation metric.

The results show that PD$^3$ also achieves competitive performance on non-power-domain datasets. This observation further reflects the adaptability of PD$^3$. Although the gain compared to the sub-optimal method may be small, it still maintains a lead across different task types, demonstrating the generalization ability of PD$^3$. We also note that, due to fundamental differences between this Question–Answer (Q–A) dataset and our task, we primarily selected bigPatent for our main experiments. Q–A datasets aim to retrieve passages that support answering a question, whereas our task is closer to Answer–Answer (A–A) similarity recognition, which focuses on comparing structurally and semantically similar texts. This distinction between Q–A and A–A tasks has been recognized in prior studies~\citep{neelakantan2022text, tseng2020effective}.

\section{Hyperparameter Analysis}
\label{section-exp-hyperparameter}

\begin{figure}[h]
  \centering
  \includegraphics[width=\linewidth]{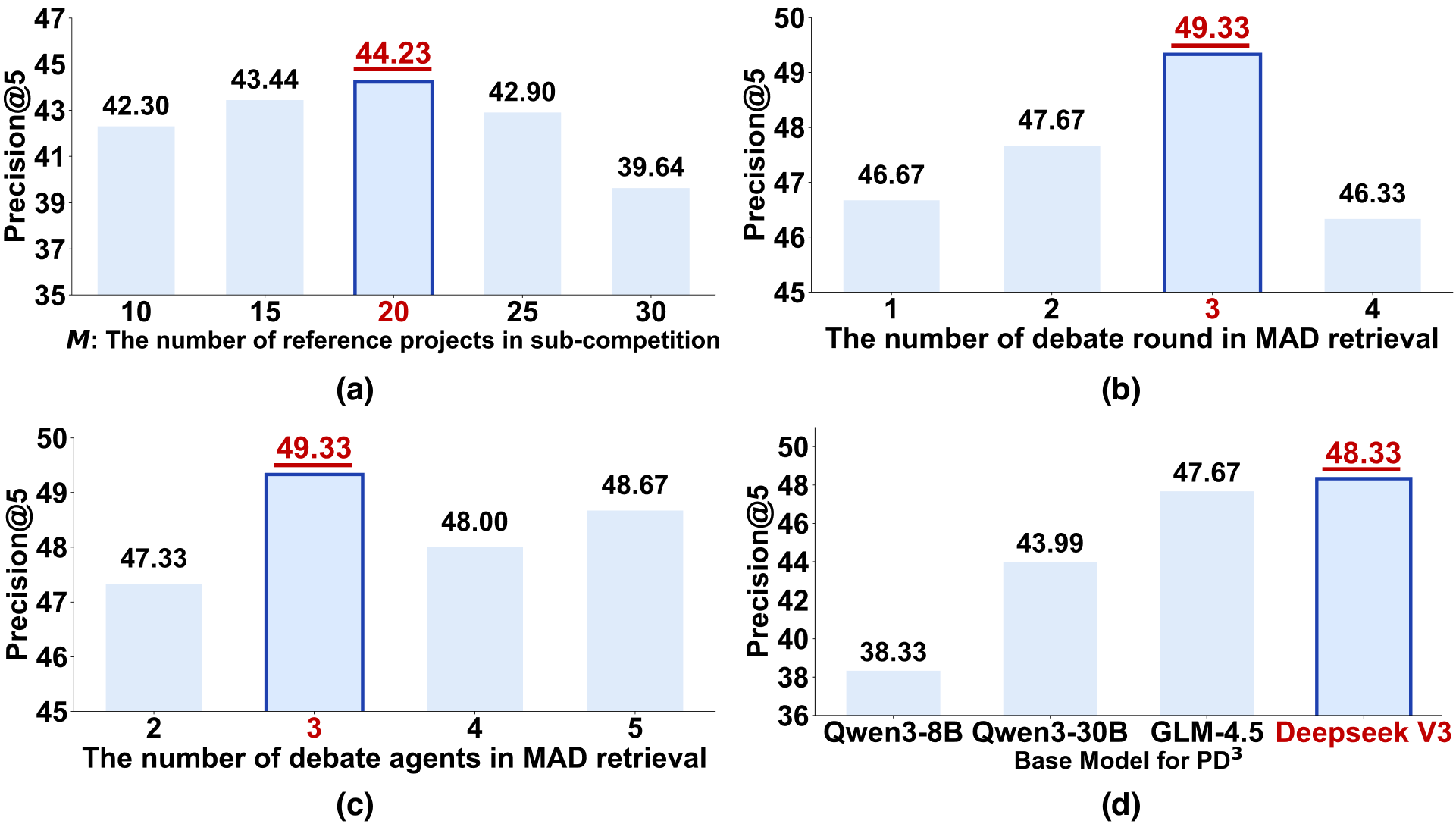}
  \caption{Hyperparameter analysis results. (a), (b), (c) and (d) present the performance comparison across hyperparameters including: group size, number of debate rounds, debate agents and base models, respectively.}
  \label{figure-hyperparameter}
\end{figure}

\vpara{Hyperparameter Experiments Settings}  Building on prior findings regarding the parameter sensitivity of MAD~\citep{smit2024should}, we are also interested in how sensitive PD$^3$'s performance is to its hyperparameters. Therefore, we conduct systematic experiments on four core hyperparameters of the MAD framework: the number of candidate items $M$ from preliminary retrieval, debate rounds, agent counts and agents' base models. 

\vpara{Hyperparameter Analysis (RQ5)} As illustrated in Figure~\ref{figure-hyperparameter} (a), experimental results across varying values of $M$ demonstrate a characteristic ``peak--then--decline" trend in reasoning performance, with optimal performance achieved at $M=20$. This phenomenon suggests that while increasing $M$ enhances the agents' access to more global information, excessively large values introduce longer context, ultimately compromising reasoning quality. 

To optimize computational costs, we conduct the following experiments on a random subset of 60 samples. As shown in Figure~\ref{figure-hyperparameter} (b), when fixing the number of agents at 3, model performance improves with additional debate rounds but declines beyond 3 rounds. Figure~\ref{figure-hyperparameter} (c) reveals that the optimal performance is achieved with 3 agents and 3 debate rounds. These findings highlight two competing factors: (1) Sufficient debate rounds and agents facilitate multi-perspective analysis and comprehensive reasoning. (2) Excessive rounds or agents introduce diminishing returns due to error propagation and context overload~\citep{estornell2024multi}. 

For the base-model study, we additionally test PD$^3$ on Qwen3-8B, Qwen3-VL-30B-A3B-Instruct~\citep{qwen3technicalreport} and GLM-4.5~\citep{zeng2025glm}, for their representative characteristics at different parameter scales. Figure~\ref{figure-hyperparameter} (d) shows that PD$^3$ indeed shows reasonable scalability across LLMs of different sizes, with performance gradually improving as the parameter size increases. At the same time, it also shows quite good performance on GLM-4.5, which is comparable to DeepSeek V3 on common benchmarks, suggesting that PD$^3$ is reasonably adaptable for deployment. See detailed hyperparameter analysis results in Table~\ref{tab:m_hyperparameter}, Table~\ref{tab:round_hyperparameter} and Table~\ref{tab:agent_hyperparameter}.

While our PD$^3$ implementation in Table~\ref{retrieval-exp-results} does not adopt the empirically optimal configuration (3 agents + 3 rounds, compared to our 3 agents + 2 rounds setting) due to cost constraints, we also evaluate PD$^3$ using the best hyperparameter configuration on the full dataset. The experimental results show that PD$^3$ further achieves 44.53\% in the Prec@5 metric, representing an improvement of 0.3\% (see Table~\ref{retrieval-exp-results-other}). Meanwhile, this also suggests that it is worthwhile to balance effectiveness and computational cost in practical applications. 

Our comprehensive evaluation of all baseline methods mentioned in Section~\ref{section-exp-setting} on subsets reveals that PD$^3$ consistently maintains superior performance, even under the worst hyperparameter configurations. As evidenced by the quantitative results in Table~\ref{tab:prec5-on-small-datasets}, this empirical observation demonstrates the framework's remarkable robustness against parameter variations. The sustained performance advantage suggests that PD$^3$'s architectural design inherently compensates for non-ideal parameter settings, a characteristic particularly valuable for real-world deployment where optimal parameter tuning may not always be feasible.

\section{Case Study and Runtime Performance}
\label{section-case-study}
\label{section-exp-runningtime}

\vpara{Case Study} \textit{Review Dingdang}'s workflow starts by retrieving 30 candidate reference projects based on vector distance. Following the round-robin competition format, the 30 candidate projects are scheduled into 15 parallel comparison groups. Figure~\ref{figure-case-study} (b) details one such comparison group: three expert agents first independently select their top-5 choices, then debate until the specified number of rounds is reached. In this case, Expert Agent A ultimately accepts the suggestions of Expert Agents B and C because their arguments are more convincing. Then the senior expert makes the final decision. After all comparison groups are completed, a voting mechanism determines the final top-5 projects. In the feedback stage, specialized LLM-as-a-Judge agents then produce both quantitative scores and qualitative assessments to the human expert.

\vpara{Runtime Performance of \textit{Review Dingdang}} To quantitatively assess the operational efficiency of the deployed system, we conduct performance evaluation focusing on three key metrics: (1) average execution time, (2) computational cost, and (3) token consumption during the detection process. We execute the detection process in parallel using Python’s `concurrent` module and Volcengine’s Model API~\citep{volcengine}, with input cost at \$0.28 per million tokens and output cost at \$1.14 per million tokens for DeepSeek V3. Our results demonstrate an average execution time of \textbf{3 minutes} per project duplication detection. As detailed in Table~\ref{tab:token_consumption}, token consumption averages 4.42 million tokens per project (4.36 million prompt tokens/57.98 thousand output tokens) during debates, while the feedback stage requires 35.36 thousand tokens per project (31.75 thousand prompt tokens/3.61 thousand output tokens). Based on Volcengine's Model API pricing~\citep{volcengine}, the system costs around \textbf{1.34 USD} per project. These metrics quantify the temporal and computational cost of PD$^3$ framework and \textit{Review Dingdang}. Compared with other baselines, PD$^3$ achieves a runtime comparable to MAD Vanilla and DMAD through its parallelization design. Although PD$^3$ involves more inference steps, the additional cost is about 1.20 USD per project. We believe this is a reasonable trade-off, given the substantial improvement in accuracy and the high financial stakes of each project. Overall, PD$^3$ provides a practically cost-effective solution for improving project duplication detection.

\begin{table}
    \centering
    \caption{Runtime and token-consumption analysis.}
    \label{tab:token_consumption}
    \resizebox{1\linewidth}{!}{
        \begin{tabular}{@{}ccrrr@{}}
        \toprule
        \textbf{Process} & \textbf{Token Consumption}  & \multicolumn{3}{c}{\textbf{Average Token Usage}}  \\ 
        \cmidrule(l){3-5}
         \textbf{Stage} & \textbf{Metrics} & input & output & total  \\ 
        \midrule
        \multirow{2}{*}{\textbf{Debate}} & per Debate & 290k  & 4k  & 294k  \\ 
        \cmidrule(l){2-5} 
         & per Project  & 4,358k   & 58k        & 4,416k \\ 
        \cmidrule{1-5}
        \textbf{Feedback}   & per Project  & 32k & 3k & 35k \\
        \cmidrule{1-5}
        \textbf{Total}   & per Project  & 4,390k & 61k & 4,451k \\
        \bottomrule
        \end{tabular}
    }
\end{table}

\vpara{Test-time scaling and economic significance.} For transparency, we report the number of LLM inference calls used by each method in the retrieval stage under our evaluation protocol. A single-LLM selection method uses one call per project, LLM-as-a-Judge uses 30 calls, Vanilla MAD uses 10 calls, LLM voting uses 15 calls, MAD Traversal uses 50 calls, and MAD Random and MAD Sliding each use 150 calls. PD$^3$ also uses 150 calls, corresponding to 15 comparison contexts with 10 calls per context. Thus, compared with a single-call method, PD$^3$ introduces a substantially larger test-time call budget. However, the economic relevance of this additional inference should be evaluated against the financial exposure of the screening task rather than by call count alone.

In the deployment of \textit{Review Dingdang}, the measured inference cost is only \textbf{1.34 USD} per project, so processing all 442 projects costs approximately 592 USD. The platform identified 41 highly duplicative projects associated with an estimated 13.44 million USD in avoided duplicate investment, corresponding to an average of approximately 327,800 USD per detected duplicate project. In other words, the estimated avoided allocation associated with one detected duplicate is equivalent to the inference cost of roughly 244,000 project screenings. Equivalently, the recorded deployment totals correspond to approximately 22,700 USD of avoided duplicate investment per inference dollar. In the live detection tests, the platform identified 24 additional highly duplicative projects associated with 8.09 million USD in avoided investment, while the full 442-project screening process still cost only about 592 USD. These figures are deployment accounting indicators rather than a causal ROI estimate, but they show that the additional test-time scaling can be economically justified: a small increase in per-project inference cost is exchanged for a materially higher probability of detecting costly duplicate proposals.

\vpara{Practical meaning of the performance gains.} The improvements in Precision@5 should be interpreted at the level of reference evidence rather than as an abstract percentage difference. Because five references are returned for each project, a one-percentage-point increase in Precision@5 corresponds to 0.05 additional expert-aligned references per project, or one additional aligned reference for every 20 reviewed projects. On the 331-case evaluation set, one percentage point corresponds to approximately 16.6 additional correct reference selections. Therefore, the 4.05-percentage-point improvement over the strongest baseline corresponds to approximately 0.2025 additional aligned references per project and about 67 additional correct reference selections across the evaluation cases. This effect becomes substantially larger at the scale of real-world funding review, where hundreds or thousands of projects may be screened each year. For example, for 1,000 projects, every one-percentage-point improvement would yield approximately 50 additional expert-aligned reference selections, while the observed 4.05-percentage-point improvement would yield approximately 203 additional selections. These additional references can provide independent evidence across research content, key technologies, and application scenarios, which may determine whether a borderline proposal receives further duplication review.

The End-to-End accuracy improvement provides a similar interpretation: the 9.77-percentage-point gain corresponds to approximately 25 additional correct pairwise decisions over the 256 test pairs. Moreover, the Match@K results show that the improvement is not limited to cases with one obvious matching project. Match@1 can be satisfied by recovering only one expert-selected reference, whereas Match@5 requires recovering the complete five-project reference set and is therefore a substantially stricter test. PD$^3$ recovers at least one expert-selected reference in 310 of 331 cases (93.66\%) and recovers all five references in 10 cases (3.02\%), while the strongest baseline recovers all five in at most 4 cases (1.21\%). PD$^3$ also achieves the best or near-best results for the intermediate values of $K$. These high-$K$ results are particularly relevant to real-world duplication detection because difficult proposals often overlap with historical work across several complementary aspects rather than having one clearly nearest neighbor. Such proposals are more likely to evade one-shot or ranking-oriented screening, whereas PD$^3$ is better able to surface a complete, evidence-bearing reference set. Thus, even a seemingly modest percentage-point improvement can translate into materially better identification of difficult-to-detect duplicate projects at deployment scale.

\section{Experiment settings details}
\label{section-appendix-expsettings} 

\vpara{Baselines.} For a comprehensive comparison, we evaluate methods from four categories: 

\vpara{Word frequency-based (WF)}: 

\underline{ROUGE-L}: Measures text similarity via longest common subsequence. \underline{BM25}: Enhanced TF-IDF approach using term frequency and inverse document frequency.

\vpara{Vector distance-based (VD)}:

\underline{gte-1.5B} (gte-Qwen2-1.5B-instruction): Transformer-based embedder. \underline{Reranker}: Larger embedders (e.g., \underline{gte-7B} (gte-Qwen2-7B-instruction), \underline{Qwen3-8B} (Qwen3-Embedding-8B) or specifically pretrained rerankers (\underline{jina} (jina-reranker-v2-base-multilingual), \underline{bge} (bge-reranker-v2-m3)).

\vpara{LLM-based (LLM)}: 

\underline{DeepSeek V3}: Open-source LLM generating responses directly from input. \underline{DeepSeek R1}: Reasoning-enhanced LLM with self critique thinking process. \underline{LLM-as-a-Judge}: Prompt LLM as a judge and execute evaluation through generation or point-wise scoring. \underline{TourRank}: LLM-based information retrieval methods with tournament format.

\vpara{MAD-based (MAD)}: 

\underline{MAD Vanilla}: Directly apply vanilla MAD for $K$-out-of-$N$ retrieval. \underline{DMAD}: Improved MAD achieving state-of-the-art performance by combining diverse reasoning approaches.

\vpara{General experiment settings} We choose DeepSeek-V3-250324 as the base model for PD$^3$ and all LLM-associated methods. As an open-source model, it provides strong Chinese and English support and competitive general benchmark performance~\citep{wang2024mmlu}. In MAD round-robin retrieval, we set the number of agents and debate rounds (excluding the initial independent selection round) to 3 and 2, following the settings of~\citep{du2023improving}. To keep the comparison fair, we use the same pre-randomly shuffled reference-project order, prompts, base-model hyperparameters, and decoding settings across baselines and hyperparameter experiments. Considering the high cost of evaluating long project proposals with multi-agent debate, we report single-run results under this fixed protocol, following common practice in prior LLM and MAD evaluations~\citep{chan2023chateval, fan2025imad, liu2025breaking, deepseekai2024deepseekv3technicalreport, zeng2025glm}. To strengthen the evidence beyond fixed decoding choices such as setting the sampling temperature to 0, we further conduct paired statistical significance tests over evaluation instances. These tests directly assess whether the observed gains are reliable across samples, providing more convincing support for the comparison than decoding determinism alone.

\vpara{Round-robin hyperparameter rationale.} Unless otherwise stated, the scheduler uses $N=30$, $M=20$, and $B=5$. Here, $N$ denotes the candidate-pool size after preliminary retrieval rather than the full reference corpus size. We choose $N=30$ as a budget-conscious high-recall operating point: the preliminary pool covers 89.40\% of human-annotated top-5 references, while increasing $N$ further raises the cost of subsequent comparisons through additional comparison contexts or longer context windows (Appendix~\ref{section-appendix-preliminary-retrieval}). We set $M=20$ to balance global candidate information against LLM context length; the dedicated hyperparameter study identifies $M=20$ as the best-performing group size (Section~\ref{section-exp-hyperparameter} and Table~\ref{tab:m_hyperparameter}). We set $B=5$ to retain fine-grained block granularity while keeping block-level enumeration practical. Together, these values give $J=N/B=6$, $r=M/B=4$, and $G=\binom{J}{r}=15$. For larger candidate pools, $N$ can be increased as needed while jointly tuning $B$ and $M$ to keep $J$, $r$, and $G$ within the deployment budget. If $N$ is not divisible by $B$, a near-balanced remainder block or padded tail block preserves the same construction with only bounded imbalance.

\vpara{Experiment settings details on Task 1} For the methods based on word frequency, vector distance retrieval, and LLM-as-a-Judge, we directly calculate the scores of the projects under detection and the preliminary retrieval results and then select the top-5 results with the highest scores. For DeepSeek V3 and DeepSeek R1, we directly input the project under detection with all preliminary retrieval results as prompts to generate outputs. For TourRank, we follow the original text setting, sequentially adopt a 30 $\rightarrow$ 20 $\rightarrow$ 10 $\rightarrow$ 5 $\rightarrow$ 2 filtering process, and run 10 times in parallel per case to determine the final result. For DMAD, we replace some of the original reasoning methods (limited to closed-ended questions) with Chain-of-Thought~\citep{wei2022chain}, Step-Back Prompting~\citep{zheng2023take} and Self-Refine~\citep{madaan2023self, kim2023language}, fixing all other experimental settings.

\vpara{Experiment settings details on Task 2} For all LLM-as-a-Judge methods (including LLM-as-a-Judge MAX, LLM-as-a-Judge AVG, and PD$^3$'s LLM-as-a-Judge-based feedback), we perform three independent generations and use the average value as the final score. While we employ the average of three scores as the final evaluation metric, instances may occur where the LLM assigns identical scores for both $P_\text{u-A}$ and $P_\text{u-B}$ in the test set. For fairness, our scoring protocol differs between evaluation settings in such cases: (1) Under the standard "ACC" setting, these cases receive a score of 0. (2) Under the "Weighted ACC" setting, we assign the score corresponding to the less frequent label in the annotation (e.g., a 2:1 ratio would yield 0.33 points). This is based on our findings in the expert manual annotation: From multiple evaluation perspectives, it is difficult to make a confident decision among several randomly selected pairs. We believe that this setting is closer to the granularity of comparison methods and manual detection.

\section{Statistical test results}
\label{section-appendix-statistical-test}

\begin{table}[h]
  \caption{Paired T-test results of Task 1.}
  \label{tab:statistical_test_task1}
  \centering
  \resizebox{\linewidth}{!}{
      \begin{tabular}{crrr}
        \toprule
        \multirow{2}{*}{\textbf{Method}} & \multirow{2}{*}{\textbf{Prec@5}} & \multicolumn{2}{c}{\textbf{T Test}} \\ 
        & & \textbf{T} & \textbf{P} \\
        \midrule
        ROUGE-L \textbf{(WF)} & 23.99 & 13.9567 & $<$0.0001  \\
        BM25 \textbf{(WF)} & 28.70 & 11.6914 & $<$0.0001 \\
        \cmidrule{1-4}
        gte-1.5B \textbf{(VD)}  & 38.13 & 5.1101 & $<$0.0001  \\
        gte-1.5B with instruction \textbf{(VD)}  & 37.82 & 5.3663 & $<$0.0001  \\
        gte-7B as reranker \textbf{(VD)}  & 38.97 & 4.5791 & $<$0.0001 \\
        Qwen3-8B as reranker  \textbf{(VD)}  & \underline{40.18} & 3.6452 & $<$0.0001  \\
        jina as reranker  \textbf{(VD)}  & 27.98 & 12.1258 & $<$0.0001  \\
        \cmidrule{1-4}
        DeepSeek V3 \textbf{(LLM)} & 36.86 & 6.9665 & $<$0.0001 \\
        DeepSeek R1 \textbf{(LLM)} & 39.52 & 4.6188 & $<$0.0001 \\
        LLM-as-a-Judge \textbf{(LLM)} & 38.49 & 5.4073 & 0.0060\\
        TourRank \textbf{(LLM)} & 37.04 & 6.4669 & 0.0060\\
        \cmidrule{1-4}
        MAD Vanilla \textbf{(LLM)} & 39.64 & 5.2801 & $<$0.0001 \\
        DMAD \textbf{(LLM)} & 38.85 & 5.6491 & $<$0.0001 \\
        \cmidrule{1-4}
        \textbf{PD$^3$ MAD Round-robin (Ours)} & \textbf{44.23} & - & - \\
        \bottomrule
      \end{tabular}
    }
\end{table}

\begin{table}[h]
  \caption{McNemar's test results of Task 2 under the Controlled Setting.}
  \label{tab:statistical_test_task2_human}
  \centering
  \resizebox{\linewidth}{!}{
      \begin{tabular}{crrrr}
        \toprule
        \textbf{Method} & \textbf{ACC} &  \textbf{$\chi^2$} & \textbf{Weighted ACC} &  \textbf{$\chi^2$} \\
        \midrule
        ROUGE-L MAX \textbf{(WF)} & 55.47 & 9.6604 & 55.34 & 24.0196  \\
        ROUGE-L AVG \textbf{(WF)}  & 55.47 & 9.4815 & 56.12 & 19.6923 \\
        \cmidrule{1-5}
        BM25 MAX  \textbf{(WF)} & 56.64 & 7.4425 & 56.77 & 16.4976\\
        BM25 AVG  \textbf{(WF)} & 55.86 & 8.2137 & 55.73 & 20.6866\\
        \cmidrule{1-5}
        gte-1.5B MAX  \textbf{(VD)}  & \underline{61.33} & 4.7377 & 41.41 & 80.1253\\
        gte-1.5B AVG  \textbf{(VD)}  & 57.42 & 10.2676 & 45.31 & 55.2660\\
        \cmidrule{1-5}
        reranker MAX \textbf{(VD)}  & - & - & - & -\\
        reranker AVG  \textbf{(VD)}  & - & - & - & -\\
        \cmidrule{1-5}
        LLM-as-a-Judge MAX  \textbf{(LLM)} & 48.44 & 35.7143 & 52.73 & 62.3077\\
        LLM-as-a-Judge AVG  \textbf{(LLM)} & 58.20 & 8.5616 & \underline{57.81} & 19.2667\\
        \cmidrule{1-5}
        \textbf{PD$^3$ (Ours)} & \textbf{67.97} & - & \textbf{64.45} & -\\
        \bottomrule
      \end{tabular}
    }
\end{table}

\begin{table}[h]
  \caption{McNemar's test results of Task 2 under the End-to-End Setting.}
  \label{tab:statistical_test_task2}
  \centering
  \resizebox{\linewidth}{!}{
      \begin{tabular}{crrrr}
        \toprule
        \textbf{Method} & \textbf{ACC} &  \textbf{$\chi^2$} & \textbf{Weighted ACC} &  \textbf{$\chi^2$} \\
        \midrule
        ROUGE-L MAX \textbf{(WF)} & 53.91 & 12.5596 & 54.04 & 34.7143  \\
        ROUGE-L AVG \textbf{(WF)}  & 55.47 & 11.0000 & 55.34 & 29.8284 \\
        \cmidrule{1-5}
        BM25 MAX  \textbf{(WF)} & \underline{58.59} & 5.5310 & 57.42 & 13.5650\\
        BM25 AVG  \textbf{(WF)} & 57.42 & 6.8772 & 56.77 & 15.9292\\
        \cmidrule{1-5}
        gte-1.5B MAX  \textbf{(VD)}  & 57.03 & 14.2542 & 46.48 & 50.4465\\
        gte-1.5B AVG  \textbf{(VD)}  & 57.42 & 15.6800 & 44.53 & 61.1237\\
        \cmidrule{1-5}
        reranker MAX  \textbf{(VD)}  & 57.42 & 13.0667 & 44.53 & 61.1237\\
        reranker AVG  \textbf{(VD)}  & \underline{58.59} & 12.2549 & 48.05 & 41.8935\\
        \cmidrule{1-5}
        LLM-as-a-Judge MAX  \textbf{(LLM)} & 41.41 & 65.2192 & 47.27 & 129.1168\\
        LLM-as-a-Judge AVG  \textbf{(LLM)} & 57.03 & 17.8963 & \underline{57.68} & 32.2874\\
        \cmidrule{1-5}
        \textbf{PD$^3$ feedback (Ours)} & \textbf{68.36} & - & \textbf{64.58} & -\\
        \bottomrule
      \end{tabular}
    }
\end{table}

The results in Table~\ref{tab:statistical_test_task1}, Table~\ref{tab:statistical_test_task2} and Table~\ref{tab:statistical_test_task2_human} show that PD$^3$ significantly outperforms all baseline methods in both tasks. In Task 1, all reported $p$-values are well below 0.05, indicating a robust improvement. In Task 2, the calculated $\chi^2$ values for all comparisons exceed the critical value of 3.8415 (corresponding to a 95\% confidence level). This confirms that the performance gains of PD$^3$ in both tasks are statistically significant.

\section{Scheduler Fairness Analysis of Ablation Variants}
\label{section-appendix-scheduler-fairness}

This appendix expands the fairness comparison used in Section~\ref{section-exp-ablation}. We use the three metrics defined in Section~\ref{section-preliminary-fairness}: the exposure gap $\Delta_{\mathrm{incl}}$, the minimum pairwise co-evaluation count $\gamma_{\mathrm{pair}}$ with low-coverage count $L_{\mathrm{pair}}(\tau)$, and the maximum context overlap $o_{\max}$ with high-redundancy count $R_{\mathrm{ctx}}(q)$. Unless otherwise noted, the calculations use the main setting $N=30$, $M=20$, $B=5$, $J=6$, $r=4$, and $G=15$. Random and Sliding use the same $G=15$ as PD$^3$; Traversal follows its sequential filtering design with five 10-candidate rounds. The following analysis gives the formulas behind the concise schedule comparison in the main text.

\begin{table*}[h]
  \caption{Complete scheduler fairness comparison for ablation variants. Random values are schedule-level expectations or Monte Carlo estimates.}
  \label{tab:scheduler-fairness-appendix}
  \centering
  \scriptsize
  \resizebox{0.95\textwidth}{!}{
      \begin{tabular}{@{}lcccccc@{}}
        \toprule
        \textbf{Scheduler} & $G$ & $\Delta_{\mathrm{incl}}$ & $\gamma_{\mathrm{pair}}$ & $L_{\mathrm{pair}}(6)$ & $o_{\max}$ & $R_{\mathrm{ctx}}(15)$ \\
        \midrule
        Round-robin & 15 & 0 & 6 & 0 & 15 & 0 \\
        Random & 15 & $\mathbb{E}[\Delta]\approx7.39$ & no deterministic guarantee & $\mathbb{E}[L]\approx128.60$ & no deterministic guarantee & $\mathbb{E}[R]\approx4.06$ \\
        Unit-step circular sliding window & 15 & 10 & 0 & 204 & 19 & 50 \\
        Traversal & 5 & 1--4, path-dependent & 0, unavoidable & not static & $\le5$ & path-dependent \\
        \bottomrule
      \end{tabular}
    }
  \vspace{-0.3cm}
\end{table*}

\vpara{Round-robin} PD$^3$ partitions the 30 candidates into $J=6$ blocks of size $B=5$ and enumerates all $r=4$ block combinations, giving $G=\binom{6}{4}=15$ comparison contexts. Each candidate appears
\begin{equation}
    a_i^{\mathrm{RR}}=\binom{J-1}{r-1}=\binom{5}{3}=10,
\end{equation}
so $\Delta_{\mathrm{incl}}^{\mathrm{RR}}=0$. Candidate pairs in the same block are co-evaluated 10 times, and candidate pairs in different blocks are co-evaluated
\begin{equation}
    \binom{J-2}{r-2}=\binom{4}{2}=6
\end{equation}
times. Therefore $\gamma_{\mathrm{pair}}^{\mathrm{RR}}=6$ and $L_{\mathrm{pair}}^{\mathrm{RR}}(6)=0$. Any two round-robin contexts share at most $B(r-1)=15$ candidates, so $o_{\max}^{\mathrm{RR}}=15$ and $R_{\mathrm{ctx}}^{\mathrm{RR}}(15)=0$.

\vpara{Random} Random samples $G=15$ distinct contexts uniformly from $\Omega_M=\{C\subseteq\mathcal{V}:|C|=M\}$, where $|\Omega_M|=T=\binom{30}{20}$. For a fixed candidate,
\begin{equation}
    a_i^{\mathrm{Rand}}\sim \mathrm{Hypergeometric}\left(T,\binom{N-1}{M-1},G\right),
\end{equation}
so $\mathbb{E}[a_i^{\mathrm{Rand}}]=GM/N=10$. However, the realized exposure gap depends on the joint vector $(a_1,\ldots,a_N)$, not on one marginal distribution. Monte Carlo estimation over 100,000 random schedules gives
\begin{equation}
    \mathbb{E}[\Delta_{\mathrm{incl}}^{\mathrm{Rand}}]\approx7.39,\qquad
    \mathrm{median}(\Delta_{\mathrm{incl}}^{\mathrm{Rand}})=7.
\end{equation}
Exact exposure balance would require all 30 candidates to appear exactly 10 times, which is a joint schedule event rather than a marginal one. A simple combinatorial upper bound gives $\Pr(\Delta_{\mathrm{incl}}^{\mathrm{Rand}}=0)\le1.45\times10^{-8}$, so random is expectation-fair rather than deterministically fair.

For a fixed pair, $c_{ij}^{\mathrm{Rand}}\sim \mathrm{Hypergeometric}(T,\binom{N-2}{M-2},G)$. Using $\tau=6$, the round-robin minimum guarantee, the expected number of low-coverage pairs is
\begin{equation}
\mathbb{E}[L_{\mathrm{pair}}^{\mathrm{Rand}}(6)]
=
\binom{N}{2}
\sum_{\ell=0}^{5}
\frac{
\binom{\binom{N-2}{M-2}}{\ell}
\binom{\binom{N}{M}-\binom{N-2}{M-2}}{G-\ell}
}{
\binom{\binom{N}{M}}{G}
}
\approx128.60.
\end{equation}
For context redundancy, the overlap $X=|C_g\cap C_h|$ between two distinct random contexts satisfies
\begin{equation}
    \Pr(X=x)=
    \frac{\binom{M}{x}\binom{N-M}{M-x}}{\binom{N}{M}-1}.
\end{equation}
Thus
\begin{equation}
\mathbb{E}[R_{\mathrm{ctx}}^{\mathrm{Rand}}(15)]
=
\binom{15}{2}
\sum_{x=16}^{19}
\frac{\binom{20}{x}\binom{10}{20-x}}{\binom{30}{20}-1}
\approx4.06.
\end{equation}
Random is therefore fair in marginal expectation, but it does not give deterministic realized-schedule guarantees under a small LLM-call budget.

\vpara{Unit-step circular sliding window} To keep the same number of groups as PD$^3$, Sliding uses 15 unit-step circular windows:
\begin{equation}
    C_g^{\mathrm{SW}}=\{v_{(g+t)\bmod N}\}_{t=0}^{M-1},
    \qquad g=0,\ldots,14.
\end{equation}
This starts from $\{v_0,\ldots,v_{19}\}$, reaches $\{v_{10},\ldots,v_{29}\}$, then wraps around to $\{v_{11},\ldots,v_{29},v_0\}$ and ends at $\{v_{14},\ldots,v_{29},v_0,\ldots,v_3\}$. The resulting exposure counts are
\[
\begin{array}{c|ccccccccccc}
a_i^{\mathrm{SW}} & 5 & 6 & 7 & 8 & 9 & 10 & 11 & 12 & 13 & 14 & 15 \\
\hline
\#\text{candidates} & 6 & 2 & 2 & 2 & 2 & 2 & 2 & 2 & 2 & 2 & 6
\end{array}
\]
so $\Delta_{\mathrm{incl}}^{\mathrm{SW}}=15-5=10$. Its pairwise co-evaluation is also imbalanced: 21 candidate pairs are never co-evaluated, so $\gamma_{\mathrm{pair}}^{\mathrm{SW}}=0$. Under the round-robin threshold $\tau=6$, the low-coverage count is
\begin{equation}
    L_{\mathrm{pair}}^{\mathrm{SW}}(6)
    =
    21+21+27+33+39+63
    =
    204.
\end{equation}
For context redundancy, the maximum overlap occurs between adjacent windows:
\begin{equation}
    o_{\max}^{\mathrm{SW}}=M-1=19.
\end{equation}
The numbers of context pairs with overlaps $16$, $17$, $18$, and $19$ are $11$, $12$, $13$, and $14$, respectively. Therefore,
\begin{equation}
    R_{\mathrm{ctx}}^{\mathrm{SW}}(15)=11+12+13+14=50.
\end{equation}
Sliding therefore keeps the same number of groups as round-robin, but it introduces positional exposure imbalance, leaves some candidate pairs uncovered, and creates more redundant adjacent contexts.

\vpara{Traversal} Traversal is path-dependent because later contexts depend on winners selected in earlier contexts. In the current setting, each round evaluates $M_{\mathrm{Trav}}=10$ candidates, keeps $w=5$ winners, and adds $u=5$ new candidates until all 30 candidates are visited, giving $G_{\mathrm{Trav}}=5$. Its total exposure count is $G_{\mathrm{Trav}}M_{\mathrm{Trav}}=50$, so even the best allocation has exposure gap at least
\begin{equation}
    \left\lceil\frac{50}{30}\right\rceil
    -
    \left\lfloor\frac{50}{30}\right\rfloor
    =1.
\end{equation}
In the worst case, an initial winner survives all five rounds while a late candidate appears once, giving $\Delta_{\mathrm{incl}}^{\mathrm{Trav}}=4$. The minimum pairwise coverage is also not guaranteed: a first-round candidate eliminated immediately cannot be co-evaluated with candidates introduced in later rounds, so $\gamma_{\mathrm{pair}}^{\mathrm{Trav}}=0$ is unavoidable. Although Traversal has bounded adjacent overlap $|C_t\cap C_{t+1}|=w=5$, its fairness depends on intermediate LLM decisions rather than on a fixed scheduler design.


\section{Prompt Templates}
\label{section-appendix-prompt}

\noindent\begin{tcolorbox}[
    colback=lightgray!10!white,    
    colframe=darkgray,  
    title=Prompt template for MAD round-robin retrieval – initial round,            
    fonttitle=\bfseries,     
    arc=3pt,           
    boxrule=0.5pt,           
    breakable,
    width=1\linewidth
    ]
    You are an expert in the power domain conducting project duplication detection named \{expert\_name\}. 
    
    Please select the five projects most relevant to the project under detection based on the following detection criteria and project information. 
    
    As an independent expert, you have no preconceived biases towards the research content of each project and focus solely on determining the best choice. 
    
    \#\# Detection Objective: 
    
    Based on predefined detection criteria and discussion procedures, strictly discuss and determine five candidate projects that are most relevant to the project under detection. 
    
    \#\# Detection Criteria: 
    
    \{detection\_criteria\} 
    
    \#\# Project Under Detection Information: 
    
    \{project\_under\_detection\_info\} 
    
    \#\# Candidate Relevant Reference Project Information: 
    
    \{candidates\_project\_info\} 
    
    Please select the five projects you consider most relevant and briefly explain the reasons for your selection in a complete sentence.
    
\end{tcolorbox}

\noindent\begin{tcolorbox}[
    colback=lightgray!10!white,    
    colframe=darkgray,  
    title=Prompt template for MAD round-robin retrieval – debate round,            
    fonttitle=\bfseries,     
    arc=3pt,           
    boxrule=0.5pt,           
    breakable,
    width=1\linewidth
    ]
    You are an expert in the power domain conducting project duplication detection named  \{expert\_name\}. 
    
    You are participating in a project duplication detection debate involving fellow experts, aiming to select the five most relevant candidate projects to the project under detection from several options. Currently, it is the \{round\_num\}th round of debate. 

    I will provide you with the detection objectives, detection criteria, relevant project information, and records of previous rounds of debate. 
    
    Please make a statement based on the previous discussions. You can: respond to the opinions of other experts, present new arguments to support your choices; or adopt the opinions of other experts, modify your previous choices and explain the reasons; or also question the choices and statements of other experts. 
    
    \#\# Detection Objective: 
    
    Based on predefined detection criteria and discussion procedures, strictly discuss and determine five candidate projects that are most relevant to the project under detection. 
    
    \#\# Detection Criteria: 
    
    \{detection\_criteria\} 
    
    \#\# Project Under Detection Information: 
    
    \{project\_under\_detection\_info\} 
    
    \#\# Candidate Relevant Reference Project Information: 
    
    \{candidates\_project\_info\} 

    \#\# Records of previous debate: 
    \{debate\_records\} 

    Please make a brief statement in a complete sentence to express your views.    
\end{tcolorbox}

\noindent\begin{tcolorbox}[
    colback=lightgray!10!white,    
    colframe=darkgray,  
    title=Prompt template for MAD round-robin retrieval – senior expert,            
    fonttitle=\bfseries,     
    arc=3pt,           
    boxrule=0.5pt,           
    breakable,
    width=1\linewidth
    ]
    As a senior expert in the power domain for project duplication detection, you are responsible for organizing expert debate to select the five most relevant projects among several candidate related projects. 

    You will serve as a discussion reviewer in this debate, evaluating the experts’ debate and determining the final five selected projects. 
    
    \#\# Detection Objective: 
    
    Based on predefined detection criteria and discussion procedures, strictly discuss and determine five candidate projects that are most relevant to the project under detection. 
    
    \#\# Detection Criteria:
    
    \{detection\_criteria\} 
    
    \#\# Project Under Detection Information: 
    
    \{project\_under\_detection\_info\} 
    
    \#\# Candidate Relevant Reference Project Information:
    
    \{candidates\_project\_info\} 
    
    \#\# Records of debate: 
    
    \{debate\_records\} 
    
    Please analyze the experts' consensus, and finally, output the list of project numbers in order of relevance after [RESULT].
\end{tcolorbox}

\noindent\begin{tcolorbox}[
    colback=lightgray!10!white,    
    colframe=darkgray,  
    title=Prompt template for LLM-as-judge-based feedback – duplication scoring,            
    fonttitle=\bfseries,     
    arc=3pt,           
    boxrule=0.5pt,           
    breakable,
    width=1\linewidth
    ]
    You are an expert in the power domain conducting project duplication detection.

    For a given project under detection, you are provided with the five most relevant historical projects from the provided database, as well as the review experts’ conclusion on the relevant content of the reference projects and the project under detection. 
    
    Your task is to score the degree of duplication of the project under detection according to the detection criteria. 
    
    \#\# Detection Criteria:
    
    Scoring is on a 10-point scale: 1 is the lowest, indicating that all historical projects are basically unrelated to the project under detection; 4-6 is in the middle, indicating that multiple historical projects overlap with the project under detection in some dimension; 10 is the highest, indicating that one or more reference projects are completely identical to the project under detection. Use the full range of scores when appropriate. 
    
    When scoring, it is necessary to comprehensively consider the similarity of the candidate relevant projects in terms of research themes, core technologies, and application scenarios. 
    
    It should be noted that if one of the five reference projects is highly relevant to the project under detection, a higher score should be given. Only when all five reference projects are not sufficiently relevant should a lower score be given.
    
    It should be noted that the five reference projects provided may not be highly relevant to the project under detection. 
    
    \#\# Project Under Detection Information: 
    
    \{project\_under\_detection\_info\} 
    
    \#\# Candidate Relevant Reference Project Information: 
    
    \{candidates\_project\_info\}
    
    \#\# Conclusion of detection Experts: 
    
    \{expert\_conclusion\} 
    
    You need to provide the reasoning first, and then give the score in the form of '[RESULT]score', where score is an integer from 1 to 10.
\end{tcolorbox}

\noindent\begin{tcolorbox}[
    colback=lightgray!10!white,    
    colframe=darkgray,  
    title=Prompt template for rewriting bigPatent in cross-domain experiments,            
    fonttitle=\bfseries,     
    arc=3pt,           
    boxrule=0.5pt,           
    breakable,
    width=1\linewidth
    ]
    You are a staff member of a patent office, and your task is to write a new patent abstract based on the given original patent. 

    The requirement is to rewrite it based on the original patent, replacing key technologies, polishing the target issue, or changing the application scenario, so that the new patent is different from the original patent and can avoid being flagged as a duplicate. You are encouraged to make substantial text modifications.
    
    Please directly output the new patent abstract; do not include any other content or markings.
    
    Original Patent Abstract: 
    \{origin\_patent\}
\end{tcolorbox}

\section{Detailed experiment results}
\label{section-appendix-expresults}

\vpara{Experiment on Task 1 results details} For vector distance-based methods, we conduct additional experiments beyond those reported in the main text. Due to space limitations, Table~\ref{retrieval-exp-results} presents only the top-performing method from each category, while Table~\ref{retrieval-exp-results-other} in the appendix provides complete experimental results. Within each category, the bold entries indicate the methods selected for Table~\ref{retrieval-exp-results} based on superior performance. Table~\ref{retrieval-exp-results-other} also presents the performance of PD$^3$ under the best hyperparameter configuration (3 agents + 3 debate rounds).

Table~\ref{tab:prec5-on-small-datasets} further shows the performance of all methods on small-scale datasets used for hyperparameter analysis. These subsets are described in Appendix~\ref{section-exp-hyperparameter}. Combining the results of hyperparameter analysis, PD$^3$ outperforms other baseline methods under different hyperparameter settings, further indicating the robustness.

\vpara{Experiment on ablation results details} Table~\ref{tab:ablation} shows the detailed results of ablation analysis. 

\vpara{Experiment on cross-domain dataset results details} Table~\ref{tab:cross-domain} shows the detailed experimental results of cross-domain analysis.

\vpara{Experiment on hyperparameters results details} Table~\ref{tab:m_hyperparameter}, Table~\ref{tab:round_hyperparameter}, and Table~\ref{tab:agent_hyperparameter} respectively show the detailed experimental results of the analysis for three key hyperparameters in PD$^3$ - the number of candidates in comparison context $M$, the number of debate rounds, and the agent count.

\begin{table*}
  \caption{Additional experiment results of Task 1.}
  \label{retrieval-exp-results-other}
  \centering
  \resizebox{0.7\textwidth}{!}{
      \begin{tabular}{crrrrrr}
        \toprule
        \multirow{3}{*}{\textbf{Method}} & \multirow{3}{*}{\textbf{Prec@5}}  & \multicolumn{5}{c}{\textbf{Match@K}} \\
        \cmidrule{3-7}
        &  & \textbf{K = 1} & \multirow{2}{*}{\textbf{K = 2}} & \multirow{2}{*}{\textbf{K = 3}} & \multirow{2}{*}{\textbf{K = 4}} & \multirow{2}{*}{\textbf{K = 5}}\\
        & & \textbf{(Hit Rate@5)} & & & &  \\
        \midrule
        \textbf{RETSim} & 27.73 & 13 | 0.0393 & 2 | 0.0060 & 0 | 0.0000 & 0 | 0.0000 & 0 | 0.0000  \\
        \cmidrule{1-7}
        gte-7B as reranker+ Task INST & 37.70 & 293 | 88.52 & 210 | 63.44 & 93 | 28.10 & 26 | 7.85 &  \textbf{2 | 0.60} \\
        \textbf{gte-7B as reranker(w/o INST)} & \textbf{38.97} & \textbf{298 | 90.03} & \textbf{212 | 64.05} & \textbf{105 | 31.72} & \textbf{29 | 8.76} & 1 | 0.30  \\
        \cmidrule{1-7}
        gte-1.5B+ EN default INST &  36.92 & 290 | 87.61 & 209 | 63.14 & 92 | 27.79 & 19 | 5.74 & 1 | 0.30  \\
        gte-1.5B+ CN default INST & 37.22 & 295 | 89.12 & 207 | 62.54  & 94 | 28.40 & 19 | 5.74 & 1 | 0.30  \\
        \textbf{gte-1.5B with instruction (+ Task INST)}  & \textbf{37.82} & \textbf{294 | 88.82} & \textbf{213 | 64.35} & \textbf{96 | 29.00} & \textbf{21 | 6.36} & \textbf{2 | 0.60} \\
        \cmidrule{1-7}
        bge as reranker  & 24.35  & 260 | 78.55 & 113 | 34.13 & 25 | 7.55 & \textbf{4 | 1.21} &  1 | 0.30  \\
        \textbf{jina as reranker}  & \textbf{27.98} & \textbf{275 | 83.08} & \textbf{141 | 42.60} & \textbf{42 | 12.69} & 1 | 0.30 & \textbf{1 | 0.30}  \\
        \cmidrule{1-7}
        \textbf{PD$^3$ (Best Hyperparameters)}  & \textbf{44.53}  & \textbf{313 | 94.56} & \textbf{238 | 71.90} & \textbf{133 | 40.18} & \textbf{47 | 14.20} &  \textbf{6 | 1.81}  \\
        \bottomrule
      \end{tabular}
    }
\end{table*}

\begin{table*}
  \caption{Additional experiment results on a subset of Task 1.}
  \label{tab:prec5-on-small-datasets}
  \centering
  \resizebox{0.4\textwidth}{!}{
      \begin{tabular}{cc}
        \toprule
        \textbf{Method} & \textbf{Prec@5} \\
        \midrule
        Random & 18.00 \\
        \cmidrule{1-2}
        ROUGE-L & 23.00 \\
        BM25 & 33.00 \\
        \cmidrule{1-2}
        gte-1.5B & 37.33  \\
        gte-1.5B with instruction & 39.33  \\
        gte-7B as reranker & 39.67\\
        Qwen3-8B as reranker & 35.00 \\
        jina as reranker & 28.33 \\
        gte-7B as reranker+ Task INST & 40.33 \\
        gte-7B as reranker(w/o INST) & 39.67 \\
        gte-1.5B+ EN default INST & 37.33 \\
        gte-1.5B+ CN default INST & 39.67  \\
        gte-1.5B with instruction (+ Task INST)  & 39.33 \\
        bge as reranker  & 26.67  \\
        jina as reranker  & 28.33  \\
        \cmidrule{1-2}
        DeepSeek V3  & 37.00 \\
        DeepSeek R1  & 42.00  \\
        LLM-as-a-Judge & 40.67 \\
        \cmidrule{1-2}
        DeepSeek V3 Voting & 45.67 \\
        \cmidrule{1-2}
        MAD Vanilla  & 43.00  \\
        DMAD & 41.00 \\
        MAD Traversal & 43.33\\
        MAD Random  &  45.33 \\
        MAD Sliding Window & \underline{46.00} \\
        \cmidrule{1-2}
        \textbf{PD$^3$ (Worst hyperparameter)}  & \textbf{46.33}  \\
        \textbf{PD$^3$ (Ours)}  & \textbf{47.67}  \\
        \textbf{PD$^3$ (Optimal hyperparameter)}  & \textbf{49.33}  \\
        \bottomrule
      \end{tabular}
    }
\end{table*}

\begin{table*}
  \caption{Ablation experiment results.}
  \label{tab:ablation}
  \centering
  \resizebox{0.8\textwidth}{!}{
      \begin{tabular}{crrrrrr}
        \toprule
        \multirow{3}{*}{\textbf{Method}} & \multirow{3}{*}{\textbf{Prec@5}}  & \multicolumn{5}{c}{\textbf{Match@K}} \\
        \cmidrule{3-7}
        &  & \textbf{K = 1} & \multirow{2}{*}{\textbf{K = 2}} & \multirow{2}{*}{\textbf{K = 3}} & \multirow{2}{*}{\textbf{K = 4}} & \multirow{2}{*}{\textbf{K = 5}}\\
        & & \textbf{(Hit Rate@5)} & & & &  \\
        \midrule
        gte-1.5B \textbf{(VD)} & \multirow{2}{*}{38.13} & \multirow{2}{*}{296 | 89.43} & \multirow{2}{*}{219 | 66.16} & \multirow{2}{*}{97 | 29.31} & \multirow{2}{*}{18 | 5.44} &  \multirow{2}{*}{1 | 0.30} \\
        (w/o MAD, w/o round-robin) & &  & & & &  \\
        \cmidrule{1-7}
        DeepSeek V3 Voting \textbf{(LLM with Voting)} \textbf{(LLM)} & \multirow{2}{*}{42.05} & \multirow{2}{*}{306 | 92.45} & \multirow{2}{*}{224 | 67.67} & \multirow{2}{*}{122 | 36.86} & \multirow{2}{*}{38 | 11.48} &  \multirow{2}{*}{\underline{6 | 1.81}} \\
        (w/o MAD, with round-robin) & &  & & & &  \\
        \cmidrule{1-7}
        MAD Vanilla \textbf{(MAD Vanilla)} & \multirow{2}{*}{39.64} & \multirow{2}{*}{307 | 92.75} & \multirow{2}{*}{229 | 69.18} & \multirow{2}{*}{\underline{129 | 38.97}} & \multirow{2}{*}{\textbf{42 | 12.69}} &  \multirow{2}{*}{3 | 0.91} \\
        (with MAD, w/o round-robin) & &  & & & &  \\
        \cmidrule{1-7}
        MAD Traversal \textbf{(Traversal)} & \multirow{2}{*}{42.11} & \multirow{2}{*}{304 | 91.84} & \multirow{2}{*}{224 | 67.27} & \multirow{2}{*}{125 | 37.76} & \multirow{2}{*}{40 | 12.08} &  \multirow{2}{*}{4 | 1.21} \\
        (with MAD, w/o round-robin) & &  & & & &  \\
        \cmidrule{1-7}
        MAD Random \textbf{(Random)} & \multirow{2}{*}{\underline{42.72}} & \multirow{2}{*}{\underline{309 | 93.35}} & \multirow{2}{*}{\underline{234 | 70.69}} & \multirow{2}{*}{125 | 37.76} & \multirow{2}{*}{35 | 10.57} &  \multirow{2}{*}{4 | 1.21} \\
        (with MAD, w/o round-robin) & &  & & & &  \\
        \cmidrule{1-7}
        MAD Sliding Window \textbf{(Sliding)} & \multirow{2}{*}{40.36} & \multirow{2}{*}{303 | 91.54} & \multirow{2}{*}{222 | 67.07} & \multirow{2}{*}{108 | 32.63} & \multirow{2}{*}{32 | 9.67} &  \multirow{2}{*}{3 | 0.91} \\
        (with MAD, w/o round-robin) & &  & & & &  \\
        \cmidrule{1-7}
        \textbf{PD$^3$ (ours)} & \multirow{2}{*}{\textbf{44.23}} & \multirow{2}{*}{\textbf{310 | 93.66}} & \multirow{2}{*}{\textbf{238 | 71.90}} & \multirow{2}{*}{\textbf{133 | 40.18}} & \multirow{2}{*}{\underline{41 | 12.39}} &  \multirow{2}{*}{\textbf{10 | 3.02}} \\
        (with MAD, with Round-robin) & &  & & & &  \\
        \bottomrule
      \end{tabular}
    }
\end{table*}

\begin{table*}
  \caption{Cross-domain experiment results on bigPatent.}
  \label{tab:cross-domain}
  \centering
  \resizebox{0.8\textwidth}{!}{
      \begin{tabular}{crrrrrr}
        \toprule
        \multirow{3}{*}{\textbf{Method}} & \multirow{3}{*}{\textbf{Prec@5}}  & \multicolumn{5}{c}{\textbf{Match@K}} \\
        \cmidrule{3-7}
        &  & \textbf{K = 1} & \multirow{2}{*}{\textbf{K = 2}} & \multirow{2}{*}{\textbf{K = 3}} & \multirow{2}{*}{\textbf{K = 4}} & \multirow{2}{*}{\textbf{K = 5}}\\
        & & \textbf{(Hit Rate@5)} & & & &  \\
        \midrule
        Random \textbf{(Random)}  & 18.80 & 67 | 67.00 & 25 | 25.00 & 2 | 2.00 & 0 | 0.00 & 0 | 0.00 \\
        \cmidrule{1-7}
        ROUGE-L \textbf{(WF)} & 43.60  & 92 | 92.00 & 72 | 72.00 & 39 | 39.00 & 13 | 13.00 & 2 | 2.00  \\
        BM25 \textbf{(WF)}  & 44.40 & 91 | 91.00 & 73 | 73.00 & 41 | 41.00 & 15 | 15.00 & 2 | 2.00 \\
        \cmidrule{1-7}
        gte-1.5B \textbf{(VD)} & 52.00 & \underline{94 | 94.00}  & 77 | 77.00 & 56 | 56.00 & 28 | 28.00 & 5 | 5.00  \\
        gte-1.5B with instruction \textbf{(VD)} & 47.20 & 92 | 92.00 & 73 | 73.00 & 48 | 48.00 & 21 | 21.00 & 2 | 2.00  \\
        gte-7B as reranker \textbf{(VD)} & 22.60  & 74 | 74.00 & 30 | 30.00 & 8 | 8.00 & 1 | 1.00 & 0 | 0.00  \\
        Qwen3-8B as reranker \textbf{(VD)} & 43.80  & 92 | 92.00 & 68 | 68.00 & 46 | 46.00 & 11 | 11.00 & 2 | 2.00  \\
        jina as reranker \textbf{(VD)} & 19.40  & 57 | 57.00 & 26 | 26.00 & 11 | 11.00 & 3 | 3.00  & 0 | 0.00  \\
        \cmidrule{1-7}
        DeepSeek V3 \textbf{(LLM)} &  76.00 & \textbf{100 | 100.00} & \underline{98 | 98.00} & 91 | 91.00 & 68 | 68.00 & 23 | 23.00  \\
        DeepSeek R1 \textbf{(LLM)} & 80.00 & \textbf{100 | 100.00} & \underline{98 | 98.00}  & \underline{97 | 97.00} & 79 | 79.00 & 26 | 26.00 \\
        LLM-as-a-Judge \textbf{(LLM)} & \underline{80.60}  & \textbf{100 | 100.00} & \textbf{99 | 99.00} & \underline{97 | 97.00} & \underline{80 | 80.00} & 27 | 27.00 \\
        TourRank \textbf{(LLM)} & 76.80  & \textbf{100 | 100.00} & \textbf{99 | 99.00}  & 91 | 91.00 & 72 | 72.00 & 22 | 22.00 \\
        \cmidrule{1-7}
        MAD Vanilla \textbf{(MAD)} & 79.60  & \textbf{100 | 100.00} & \textbf{99 | 99.00} & 93 | 93.00 & 75 | 75.00 &  \underline{31 | 31.00}  \\
        DMAD \textbf{(MAD)} & 71.20  & \textbf{100 | 100.00} & \underline{98 | 98.00} & 86 | 86.00 & 57 | 57.00 &  15 | 15.00  \\
        \cmidrule{1-7}
        \textbf{PD$^3$ MAD Round-robin (Ours)}  & \textbf{83.60}  & \textbf{100 | 100.00} & \textbf{99 | 99.00}  & \textbf{98 | 98.00} & \textbf{84 | 84.00}  & \textbf{37 | 37.00}  \\
        \bottomrule
      \end{tabular}
    }
    \vspace{-0.3cm}
\end{table*}

\begin{table*}
    \caption{Hyperparameter analysis experiment results on the group size.}
    \label{tab:m_hyperparameter}
    \centering
    \scriptsize
    \resizebox{0.8\linewidth}{!}{
        \begin{tabular}{@{}crrrrrr@{}}
        \toprule
        \textbf{Round-robin} & \multirow{3}{*}{\textbf{Prec@5}} & \multicolumn{5}{c}{\textbf{Match @ K}}  \\ 
        \cmidrule(l){3-7} 
        \textbf{initial item count}  &  & \textbf{K = 1} & \multirow{2}{*}{\textbf{K = 2}} & \multirow{2}{*}{\textbf{K = 3}} & \multirow{2}{*}{\textbf{K = 4}} & \multirow{2}{*}{\textbf{K = 5}}\\
        & & \textbf{(Hit Rate@5)} & & & &  \\
        \midrule
        10   & 42.30  & 307 | 92.75    &  \underline{229 | 69.18}   & 119 | 35.95   & 39 | 11.78   & \underline{6 | 1.81}  \\
        \cmidrule{1-7}
        15 & \underline{43.44}   &  \underline{309 | 93.35} & \underline{229 | 69.18}   & \underline{132 | 39.88}   & \textbf{45 | 13.60} & 4 | 1.21    \\ 
        \cmidrule{1-7}
        \textbf{20 (ours)}    & \textbf{44.23}  & \textbf{310 | 93.66} &   \textbf{238 | 71.90}   & \textbf{133 | 40.18} & 41 | 12.39 &  \textbf{10 | 3.02}\\ 
        \cmidrule{1-7}
        25  & 42.90   & 307 | 92.75  & \underline{229 | 69.18}  & 129 | 38.97  & \underline{42 | 12.69} &  3 | 0.91    \\ 
        \cmidrule{1-7}
        30  & 39.64  & 307 | 92.75  &  \underline{229 | 69.18}  & 129 | 38.97   &   \underline{42 | 12.69}  & 3 | 0.91  \\ 
        \bottomrule
        \end{tabular}
    }
\end{table*}

\begin{table*}
    \centering
    \caption{Hyperparameter analysis experiment results on the debate round.}
    \label{tab:round_hyperparameter}
    \scriptsize
    \resizebox{0.7\textwidth}{!}{
        \begin{tabular}{@{}crrrrrr@{}}
            \toprule
            \multirow{3}{*}{\textbf{Debate rounds}} & \multirow{3}{*}{\textbf{Prec@5}}  & \multicolumn{5}{c}{\textbf{Match@K}} \\
            \cmidrule{3-7}
            &  & \textbf{K = 1} & \multirow{2}{*}{\textbf{K = 2}} & \multirow{2}{*}{\textbf{K = 3}} & \multirow{2}{*}{\textbf{K = 4}} & \multirow{2}{*}{\textbf{K = 5}}\\
            & & \textbf{(Hit Rate@5)} & & & &  \\
            \midrule
            1   & 46.67  & 56 | 93.33 &  \underline{45 | 75.00}  & 28 | 46.67   &  \underline{9 | 15.00}   & \textbf{2 | 3.33}  \\ 
        \cmidrule{1-7}
            2    & \underline{47.67}     & \underline{57 | 95.00} &  \textbf{47 | 78.33} & 29 | 48.33  & 8 | 13.33 &   \textbf{2 | 3.33}  \\ 
        \cmidrule{1-7}
            \textbf{3}  & \textbf{49.33} & \textbf{58 | 96.67}  & \underline{45 | 75.00} & \textbf{33 | 55.00} & \textbf{11 | 18.33}   & \underline{1 | 1.67} \\ 
        \cmidrule{1-7}
            4   & 46.33  &  \underline{57 | 95.00}   &  43 | 71.61 & \underline{30 | 50.00} & 8 | 13.33    & \underline{1 | 1.67}          \\ 
            \bottomrule
        \end{tabular}
    }
\end{table*}

\begin{table*}
    \centering
    \caption{Hyperparameter analysis experiment results on the number of debate agents.}
    \label{tab:agent_hyperparameter}
    \scriptsize
    \resizebox{0.7\textwidth}{!}{
    \begin{tabular}{@{}crrrrrr@{}}
        \toprule
        \multirow{3}{*}{\textbf{Agent Count}} & \multirow{3}{*}{\textbf{Prec@5}}  & \multicolumn{5}{c}{\textbf{Match@K}} \\
            \cmidrule{3-7}
            &  & \textbf{K = 1} & \multirow{2}{*}{\textbf{K = 2}} & \multirow{2}{*}{\textbf{K = 3}} & \multirow{2}{*}{\textbf{K = 4}} & \multirow{2}{*}{\textbf{K = 5}}\\
            & & \textbf{(Hit Rate@5)} & & & &  \\
        \midrule
        2   & 47.33   & 56 | 93.33 & \textbf{48 | 80.00} & 25 | 41.67   & \textbf{12 | 20.00} & \textbf{1 | 1.67}          \\ 
        \cmidrule{1-7}
        \textbf{3}   & \textbf{49.33}   & \textbf{58 | 96.67} &     45 | 75.00       & \underline{33 | 55.00}          & \underline{11 | 18.33}         & \textbf{1 | 1.67} \\ 
        \cmidrule{1-7}
        4   & \underline{48.00}    & \underline{57 | 95.00} &    \underline{46 | 76.67}       & 30 | 50.00          & 10 | 16.67          & \textbf{1 | 1.67}          \\ 
        \cmidrule{1-7}
        5    & 48.67    & 56 | 93.33 & 45 | 75.00    & \textbf{34 | 56.67} & 10 | 16.67          & \textbf{1 | 1.67}          \\ 
        \bottomrule
    \end{tabular}
    }
\end{table*}


\end{document}